\documentclass{article}
\PassOptionsToPackage{numbers,sort&compress}{natbib}
\usepackage[final,main]{neurips_2026}

\usepackage[utf8]{inputenc} 
\usepackage[T1]{fontenc}    
\usepackage{hyperref}       
\usepackage{url}            
\usepackage{booktabs}       
\usepackage{amsfonts}       
\usepackage{nicefrac}       
\usepackage{microtype}      
\usepackage[table]{xcolor}         
\usepackage{microtype}
\usepackage{graphicx}
\usepackage{subcaption}
\usepackage{booktabs} 
\usepackage{makecell}
\usepackage{threeparttable}
\usepackage{amsmath}
\usepackage{amssymb}
\usepackage{bm}
\usepackage{mathtools}
\usepackage{amsthm}
\usepackage{multirow}
\usepackage{enumitem}
\usepackage{algorithm}
\usepackage{algorithmic}
\usepackage{wrapfig}

\usepackage[capitalize,noabbrev]{cleveref}

\theoremstyle{plain}
\newtheorem{theorem}{Theorem}[section]
\newtheorem{proposition}[theorem]{Proposition}

\theoremstyle{definition}
\newtheorem{definition}[theorem]{Definition}
\newtheorem{assumption}[theorem]{Assumption}
\theoremstyle{remark}
\newtheorem{remark}[theorem]{Remark}

\usepackage[textsize=tiny]{todonotes}
\newcommand{\vpara}[1]{\vspace{1.5ex}\noindent\textbf{#1}}

\newcommand{\tb}[1]{\textbf{#1}}
\newcommand{\ul}[1]{\underline{#1}}

\title{Enabling VLA Action Self-Verification via VLM Token Probability Bucketing}

\author{
  Chen Zhao\textsuperscript{\normalfont{1,2$\ast$}},
  Zhuoran Wang\textsuperscript{\normalfont{1,2}},
  Haoyang Li\textsuperscript{\normalfont{1,2$\ast$}},
  Shifeng Bao\textsuperscript{\normalfont{1,2$\ast$}},
  Guanlin Li\textsuperscript{\normalfont{1,2$\ast$}},\\
  \textbf{Youhe Feng\textsuperscript{\normalfont{1,2$\ast$}},
  Yang Li\textsuperscript{\normalfont{1,2}},
  Jie Tang\textsuperscript{\normalfont{3}},
  Jing Zhang\textsuperscript{\normalfont{1,2$\dagger$}},}
  \\
  \textsuperscript{1} Renmin University of China, Beijing 100872, China\\
  \textsuperscript{2} Key Laboratory of Data Engineering and KnowledgeEngineering, MOE, China\\
  \textsuperscript{3} Tsinghua University\\
  \texttt{2021201645@ruc.edu.cn, zhang-jing@ruc.edu.cn}\\
}

\begin{document}

\maketitle

\begingroup\def\thefootnote{$\ast$}\footnotetext{Done as intern at Zhipu AI.}\endgroup
\begingroup\def\thefootnote{$\dagger$}\footnotetext{Corresponding author.}\endgroup

\begin{abstract}
Test-time sampling can improve Vision-Language-Action (VLA) policies, but only if the model can efficiently select a good action from multiple candidates. Prior work often relies on separately trained external verifiers, adding extra models and inference overhead. We propose \textbf{Token Probability Bucketing (TokenPB)}, a self-verification approach that discretizes continuous action chunks into bucket tokens and co-trains a single VLA checkpoint with a joint objective: a flow-matching loss for action generation and an auxiliary bucket loss $\mathcal{L}_{\text{bucket}}$ that teaches the VLM backbone to model observation-conditioned distributions over bucket-token sequences. At inference, TokenPB samples $M$ candidates, scores each via teacher-forced evaluation under the learned distribution, and selects the best---enabling efficient batched scoring with observation-prefix KV-cache reuse. We provide theory connecting the training objective to a monotonic ranking property and bounds on expected selection regret that separate candidate-set effects from calibration. Across multiple backbones and benchmarks, TokenPB yields consistent success-rate improvements over single-sample inference from the same checkpoint (+2.7 in simulation, +17.1 in real-world, +16.4 under out-of-distribution shifts), and further improves over external-verifier baselines under the same sampling budget.
\end{abstract}

\section{Introduction}
Vision-Language-Action (VLA) policies must translate rich perceptual-language context into continuous, high-dimensional robot actions. In manipulation, the same observation can admit multiple plausible low-level action modes, while small deviations in a single executed action chunk can cause irreversible failures. This motivates \emph{test-time inference with sampling-and-selection}: generate multiple candidate action chunks and execute the most promising one. In practice, however, sampling is only useful if the policy can efficiently and reliably score candidates without introducing additional models or large inference overhead.

A common approach is to train a separate external verifier to rank sampled actions. While effective, external verifiers add training complexity, require additional data or supervision choices, and often incur a forward pass per candidate at inference time. More fundamentally, action quality is defined in a continuous space, whereas the VLM backbone natively produces probabilities over discrete tokens. This mismatch makes it difficult to build a verifier that is both accurate and computationally efficient.

We propose Token Probability Bucketing (TokenPB), a self-verification framework that turns continuous-action verification into a token-probability estimation problem. TokenPB discretizes each continuous action chunk with a fixed, deterministic tokenizer into a compact sequence of bucket tokens. To endow the VLA with self-verification ability, we co-train the VLM backbone with two complementary signals within a single checkpoint: a flow-matching loss $\mathcal{L}_{\text{fm}}$ drives a diffusion-style action expert to generate continuous action chunks from VLM context features, while an auxiliary bucket loss $\mathcal{L}_{\text{bucket}}$ teaches the same backbone to model observation-conditioned distributions over bucket-token sequences. At inference time, we sample $M$ candidate action chunks by varying the injected noise, tokenize each candidate into bucket tokens, and score them via teacher-forced evaluation under the learned distribution. The highest-scoring candidate is selected for execution. Because all candidates share the same observation prefix, scoring can be implemented with batched decoding and observation-prefix KV-cache reuse, making test-time sampling practical.

\begin{figure}
  \centering
  \includegraphics[width=0.87\textwidth]{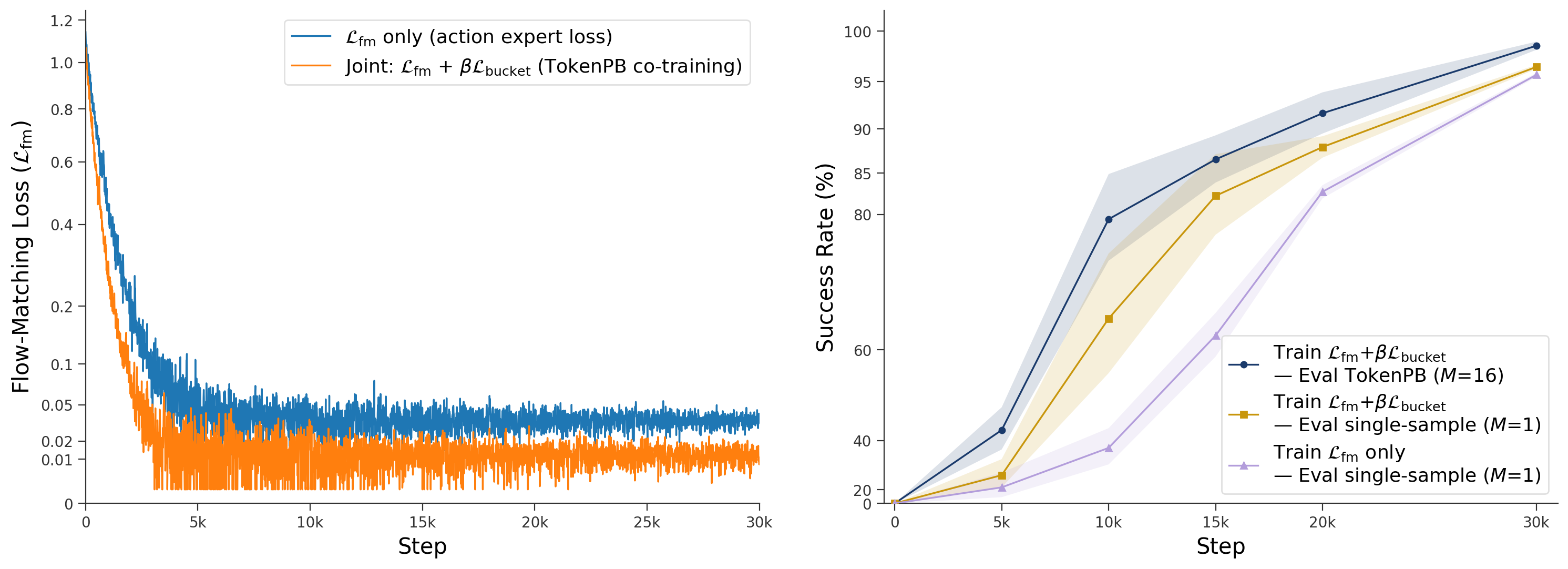}
  \caption{\textbf{Bucket co-training improves optimization and success.}
\textbf{Left:} adding $\mathcal{L}_{\text{bucket}}$ alongside the flow-matching loss $\mathcal{L}_{\text{fm}}$ accelerates convergence (lower $\mathcal{L}_{\text{fm}}$ at the same step).
\textbf{Right:} at every checkpoint, joint training ($\mathcal{L}_{\text{fm}}+\beta\,\mathcal{L}_{\text{bucket}}$) yields higher downstream success rates than the $\mathcal{L}_{\text{fm}}$-only baseline.}

  \label{fig:co_training}
\end{figure}

A natural concern with joint training is that the auxiliary bucket objective might interfere with the primary action generation task. Figure~\ref{fig:co_training} shows that the opposite is true: the bucket loss \emph{accelerates} flow-matching convergence (left panel) and, at every training checkpoint, the jointly trained model achieves higher downstream success rates than a baseline trained only with the flow-matching action expert loss $\mathcal{L}_{\text{fm}}$ (right panel). This synergy arises because the two objectives share the VLM backbone, and $\mathcal{L}_{\text{bucket}}$ provides a complementary regularization signal that sharpens action-relevant representations beneficial to both generation and scoring.

We further provide theoretical support for the bucket score as a candidate ranker. We establish a \emph{scoring monotonicity} property: as the learned distribution converges toward the true expert-induced distribution, the ranking produced by the bucket score becomes consistent with the true quality ordering among candidates. Building on this, we derive bounds on expected selection regret that separate two effects: (i) a candidate-set term capturing how the number and diversity of sampled candidates influence selection difficulty, and (ii) a calibration term measuring how well the learned distribution approximates the true one. This decomposition clarifies when test-time sampling is most beneficial and why improving calibration directly improves selection.

Our contributions are threefold: (1) we introduce TokenPB, which enables efficient test-time inference with sampling-and-selection inside a single VLA checkpoint by combining $\mathcal{L}_{\text{fm}}$ with an auxiliary $\mathcal{L}_{\text{bucket}}$ objective and performing batched, prefix-cached scoring; (2) we provide monotonicity guarantees and regret bounds that disentangle candidate-set effects from calibration; and (3) we demonstrate consistent improvements across multiple backbones and benchmarks, including significant gains from test-time inference over single-sample inference and further improvements over external-verifier baselines under the same sampling budget.

\section{Related Work}
\label{sec:related}
\vpara{External action verification.}
Prior work often improves VLA reliability by generating multiple candidate actions at test time and
using an auxiliary scorer beyond the base generator for selection.
RoboMonkey \citep{robomonkey2025} trains a dedicated VLM-based verifier to score and select among
candidates, while RoVer \citep{rover2025} learns a process reward model that provides process-level
quality estimates for ranking.
V-GPS \citep{vgps2025} instead uses an offline RL-learned value function as a plug-in critic to
re-rank actions from a frozen generalist policy at deployment.
AsyncVLA \citep{asyncvla2025} is related but different: it trains an additional confidence rater
to estimate token-level confidence and triggers selective self-correction via regeneration, rather than
sampling-and-ranking action candidates.
Overall, these approaches typically require training and deploying additional scoring modules (verifier,
reward/value model, or confidence rater) alongside the base VLA.
In contrast, TokenPB co-trains the verifier within the VLM backbone itself, enabling test-time selection
without a separately deployed external scorer.

\vpara{Verifier-free selection.} MG-Select \citep{mgselect2025} scores candidates using an autoregressive VLA's own masking distribution. Its scoring mechanism is tied to action-token generation, whereas TokenPB separates continuous action generation by the action expert from teacher-forced token scoring by the VLM. This allows the two stages to scale independently and permits reuse of the sampler's observation-prefix KV cache.

\vpara{Connection to offline RL and Q-functions.} A Q-function estimates expected cumulative future reward, whereas TokenPB learns the observation-conditioned demonstration distribution $P^*(c\mid s)$ and uses it to rank a finite set of sampled candidates. Value-based approaches such as V-GPS require a learned critic; representative offline RL algorithms include CQL~\citep{kumar2020conservative}, TD3+BC~\citep{fujimoto2021minimalist}, and IQL~\citep{kostrikov2021implicit}. TokenPB instead uses the same demonstration pairs as behavior cloning, without reward annotations, preference labels, or Bellman backups. Co-training the sampler and verifier on shared data focuses verification on the sampler's candidates, although it does not guarantee that all candidates are in distribution. Its scores express conditional demonstration preference and need not identify the globally optimal action. Appendix~\ref{app:q_comparison} compares these learning targets and evaluates the learned Q-functions under a shared-sampler protocol.

\section{Preliminaries}
\vpara{Problem Statement.}\label{sec:problem} Let $s_t$ denote the observation state at time step $t$, and let $Q(s_t)$ denote the task completion progress at state $s_t$. We define the single-step utility of an action $a$ as:
$$
R(a) = Q(s_{t+1} \mid a, s_t) - Q(s_t).
$$

Since $Q(s_t)$ does not depend on $a$, maximizing $R(a)$ is equivalent to maximizing $Q(s_{t+1}\mid a, s_t)$; we use the latter for brevity.
Suppose a VLA policy generates $M$ candidate actions via test-time sampling, yielding the candidate set $\mathcal{A}_{s_t} = \{a^{(1)}, a^{(2)}, \dots, a^{(M)}\}$. We denote the oracle-best candidate as:
$$
a^* = \arg\!\!\max_{a^{(m)} \in \mathcal{A}_{s_t}} Q(s_{t+1} \mid a^{(m)}, s_t).
$$
Note that $a^*$ is constrained to the candidate set and may differ from the global optimum over the full action space. 
The objective of the VLM-based verifier is to select the candidate $\hat{a}^*$ that best approximates $a^*$. We define the \emph{verifier effectiveness} as the utility difference between the verifier's selection and the oracle-best candidate:
$$
\Delta_{\text{verify}}(s_t) = Q(s_{t+1} \mid \hat{a}^*, s_t) - Q(s_{t+1} \mid a^*, s_t),
$$
where $\Delta_{\text{verify}} \leq 0$ by definition and $\Delta_{\text{verify}} = 0$ corresponds to perfect verification. For convenience, Section~\ref{sec:bucketing} bounds $|\Delta_{\text{verify}}(s_t)|$, which equals
$Q(s_{t+1}\mid a^*, s_t)-Q(s_{t+1}\mid \hat a^*, s_t)$ under our definition.


\vpara{Flow-Matching Action Generation.}\label{sec:flow_matching_prelim}
Modern VLA architectures adopt flow matching~\citep{lipman2022flow} to generate continuous actions from the VLM hidden representation.
Given the VLM context feature $f_c$ extracted from the current observation $s_t$,
a lightweight diffusion action expert learns a conditional velocity field
$v_\theta(x_\tau, \tau \mid f_c)$ that transports a standard Gaussian
$x_0 \sim \mathcal{N}(0, I)$ toward the expert action distribution.
Concretely, we sample an expert demonstration action chunk $a^{\text{demo}} \sim \mathcal{D}$ and consider the straight-line interpolation path
$x_\tau = (1-\tau)x_0 + \tau a^{\text{demo}} $, with $\tau \in [0,1]$,
and train $v_\theta$ by matching the constant target velocity along this path:
\begin{equation}\label{eq:flow_matching}
\mathcal{L}_{\text{fm}} = \mathbb{E}_{\tau, x_0, a^{\text{demo}}}\bigl[\|v_\theta(x_\tau, \tau \mid f_c) - (a^{\text{demo}} - x_0)\|^2\bigr].
\end{equation}
At inference time, an action chunk is produced by integrating $v_\theta$ from $\tau{=}0$ to $\tau{=}1$ using a few-step Euler solver.
Because the action expert is decoupled from the heavy VLM backbone and shares only the hidden feature $f_c$,
sampling multiple candidates incurs negligible overhead compared to a single VLM forward pass.
\section{Method}
\begin{figure}
  \centering
  \includegraphics[width=\textwidth]{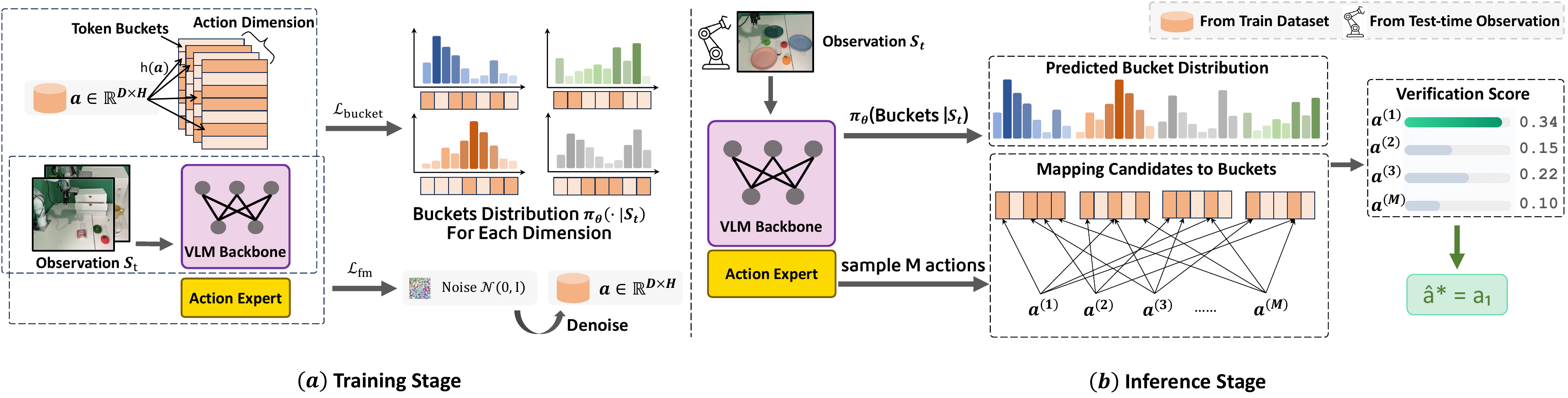}
  \caption{\textbf{Overview of the TokenPB framework.} Left: the VLM backbone learns a mapping from observation states to action token bucket distributions. Right: at test time, the VLA generates multiple candidate action chunks via the diffusion action expert, then scores each candidate by its alignment with the action token bucket probability distribution.}
  \label{fig:method_overview}
\end{figure}

TokenPB follows a \emph{sample-and-verify} paradigm (Figure~\ref{fig:method_overview}): the diffusion action expert samples multiple candidate action chunks, and the VLM backbone scores each candidate through an observation-conditioned distribution over discretized action buckets. Below we describe how actions are discretized (Section~\ref{sec:bucketing}), how candidates are sampled, scored, and co-trained (Section~\ref{sec:cotrain}), and how the resulting verifier's selection error is bounded (Section~\ref{sec:theory}).

\subsection{Action Token Bucketing}\label{sec:bucketing}

We discretize a continuous action chunk into a sequence of discrete tokens via a fixed, deterministic tokenizer $h$ as the mapping function.
For an action chunk $a\in\mathbb{R}^{D\times H}$ (action dimension $D$, chunk horizon $H$), the tokenizer outputs a token sequence
\[
c = h(a)\in\{1,\ldots,C\}^{L},
\]
where $C$ is the token vocabulary size and $L$ is the resulting sequence length\footnote{In our implementation, $h$ is instantiated by Textual FAST (Appendix~\ref{appendix:token}), which transforms action chunks into compact discrete tokens and then encodes them using the VLM tokenizer; prediction-head reuse and score normalization are detailed in Appendix~\ref{app:textualfast}.}. The mapping $h$ is non-learned and admits a deterministic decoder $g$ that reconstructs a continuous proxy action $\tilde a=g(c)$.
Due to quantization in the tokenization process, multiple continuous actions can map to the same token sequence; thus reconstruction is only defined up to the tokenizer's quantization resolution.

\paragraph{Discretization and selection.}
Tokenization introduces a quantization gap between a continuous candidate and its decoded proxy action.
Let $a^{*}=\arg\max_{a^{(m)}\in\mathcal{A}_{s_t}} Q(s_{t+1}\mid a^{(m)}, s_t)$ be the oracle-best candidate in the sampled set,
and let $\tilde a^{*}=g(h(a^{*}))$ be the decoded proxy action of its token sequence.
The total verification gap decomposes as:
\[
|\Delta_{\text{verify}}|
\le
\underbrace{\big|Q(s_{t+1}\mid \tilde a^{*}, s_t)-Q(s_{t+1}\mid \hat a^{*}, s_t)\big|}_{\text{selection error}}
+
\underbrace{\big|Q(s_{t+1}\mid a^{*}, s_t)-Q(s_{t+1}\mid \tilde a^{*}, s_t)\big|}_{\text{discretization error}}.
\]

The discretization error depends solely on the bucket granularity and is independent of the verifier.
We observe diminishing returns beyond 1024 buckets, suggesting discretization is no longer the dominant bottleneck (Figure~\ref{fig:table_charts}(right) in Section~\ref{sec:analysis}).
We therefore focus on bounding the selection error in Section~\ref{sec:theory}.

\subsection{Candidate Sampling and Verification}
\label{sec:cotrain}

\paragraph{Sample.}
As depicted in Figure~\ref{fig:method_overview}, conditioned on the VLM context feature $f_c=f_\theta(s_t)$,
the diffusion action expert generates $M$ candidate action chunks by varying the injected noise, yielding
$\mathcal{A}_{s_t}=\{a^{(1)},\ldots,a^{(M)}\}$.
Because the diffusion action  expert is lightweight, the sampling overhead is negligible relative to a VLM forward pass.

\paragraph{Verify.}
The VLM backbone defines a conditional distribution over \emph{action-token sequences} given the current observation,
denoted $\pi_\theta(\cdot\mid s_t)$.
For a candidate $a^{(m)}$, we map it into $c^{(m)}=h(a^{(m)})=(c^{(m)}_1,\ldots,c^{(m)}_L)$ and assign a score under $\pi_\theta$:
\[
\text{score}(a^{(m)}) \;=\; \frac{1}{L}\log \pi_\theta\!\big(c^{(m)}\mid s_t\big)
\;=\; \frac{1}{L}\sum_{\ell=1}^{L}\log \pi_\theta\!\big(c^{(m)}_\ell \mid s_t, c^{(m)}_{<\ell}\big).
\]
We then select
\[
\hat a^{*}=\arg\!\!\max_{\!\!a^{(m)}\in\mathcal{A}_{s_t}} \!\!\text{score}(a^{(m)}).
\]
This verification stage requires no external model: $\pi_\theta(\cdot\mid s_t)$ is computed by the same VLM backbone used for action generation.
Moreover, because all candidates share the same observation prefix, we encode the prefix once and reuse its transformer KV cache.
For each candidate $a^{(m)}$, we score only the candidate-specific token continuation in a single teacher-forced pass conditioned on the cached prefix, batching across all $m$.
This yields a small additional overhead in practice (Section~\ref{sec:analysis}), without requiring a separate verifier or an expensive per-candidate recomputation of the shared prefix.

\paragraph{Co-training objective.}
To learn the bucket-scoring distribution $\pi_\theta(\cdot\mid s)$, we jointly optimize the
flow-matching action generation loss $\mathcal{L}_{\mathrm{fm}}$ and an auxiliary \emph{token-bucket cross-entropy} on demonstration bucket-token sequences:

\[
\mathcal{L}_{\text{total}}=\mathcal{L}_{\text{fm}}+\beta\cdot \mathcal{L}_{\text{bucket}},
\qquad
\mathcal{L}_{\text{bucket}}
=-\mathbb{E}_{(s_t,a^{\text{demo}})\sim\mathcal{D}}
\Big[\frac{1}{L^{\text{demo}}}\log \pi_\theta\big(h(a^{\text{demo}})\mid s_t\big)\Big],
\]
where $L^{\text{demo}}$ is the token sequence length produced by $h$ on the demonstration action chunk. As training converges, $\pi_\theta(\cdot\mid s)$ approaches the demonstration-induced conditional distribution over token sequences.

\subsection{Selection Error Bound}\label{sec:theory}

We now bound the verifier's expected sub-optimality. 
The central question is: how much regret does the verifier incur by ranking candidates via the bucket score rather than true action quality?

The connection between the two rests on a natural assumption: candidates whose bucket representations are more frequent under expert demonstrations tend to have higher task utility. The analysis is conditional on this alignment between demonstration likelihood and task utility.


Since $\mathcal{L}_{\text{bucket}}$ is strictly proper, $\pi_\theta(\cdot\mid s_t)\to P^*(\cdot\mid s_t)$ as training proceeds,
where $P^*(\cdot\mid s_t)$ denotes the (unknown) ground-truth distribution of token sequences induced by the expert demonstrations in $\mathcal{D}$ conditioned on state $s_t$.
For our verifier, only the token sequences that appear among the sampled candidates matter.
We therefore quantify the residual mismatch via a candidate-set calibration error in log space:

\begin{equation}
\label{eq:calibration_error}
\varepsilon_{\text{cal}}
:= \mathbb{E}_{s_t}\Big[
\max_{m\in\{1,\ldots,M\}}
\big|
\log \pi_\theta(h(a^{(m)})\mid s_t)-
\log P^*(h(a^{(m)})\mid s_t)
\big|
\Big].
\end{equation}

Here $P^*(h(a^{(m)})\mid s_t)$ and $\pi_\theta(h(a^{(m)})\mid s_t)$ denote probabilities of the entire token sequence produced by $h$. Once $\varepsilon_{\mathrm{cal}}$ is sufficiently small, $\pi_\theta$ preserves the true quality ordering among candidates---a property we call \emph{scoring monotonicity}:
\begin{equation}\label{eq:scoring_monotonicity}
P^*(h(a^{(m)}) \mid s_t) > P^*(h(a^{(m')}) \mid s_t) \;\;\Longrightarrow\;\; \pi_\theta(h(a^{(m)}) \mid s_t) > \pi_\theta(h(a^{(m')}) \mid s_t).
\end{equation}
The formal statements, assumptions, and proofs are given in Appendices~\ref{app:scoring_theory} and~\ref{app:calibration_proof}.

\vpara{Expected selection error bound.} For analysis, we model the deterministic $\arg\max$ over $\text{score}(a^{(m)})$ as a Gibbs selector with effective sharpness $\lambda$---a standard surrogate that recovers the argmax in the limit $\lambda \to \infty$. The expected selection regret then satisfies:
\begin{equation}\label{eq:selection_bound}
\mathbb{E}\bigl[Q(s_{t+1} \mid a^*, s_t) - Q(s_{t+1} \mid \hat{a}^*, s_t)\bigr] \;\leq\; \underbrace{\frac{\log M}{\lambda}}_{\text{candidate diversity}} + \underbrace{O(\varepsilon_{\mathrm{cal}})}_{\text{calibration gap}}.
\end{equation}
The first term reflects the intrinsic cost of choosing among $M$ candidates with a soft scorer; the second vanishes as calibration improves. The full derivation is provided in Appendix~\ref{app:selection_error}.

\section{Experiments}
\label{sec:experiments}
\begin{table*}[t]
    \centering
    \caption{\textbf{Comparison with Single-Sample vs.\ Test-Time Inference.} All models sharing the same backbone are co-trained with both bucket-token and flow-matching objectives using a single checkpoint; configurations differ only at inference time. ``Model-GR00T'': single sample ($M{=}1$) without verification. ``Model-GR00T+TokenPB'': multi-candidate verification ($M{=}16$). For Real-World, ``Seen'' denotes in-distribution tasks and ``Unseen'' denotes held-out instructions and object configurations not seen during training. Detailed training settings are provided in Table~\ref{12}.}
    \label{table:single_vs_tti}
    \renewcommand{\arraystretch}{1.15}
    \resizebox{\linewidth}{!}{
    \begin{tabular}{@{}l l  c c c c c c@{}}
    \toprule
    \textbf{Benchmark} & \textbf{Model} & \multicolumn{6}{c}{\textbf{Success Rate (\%)}} \\
    \midrule
    \rowcolor{gray!8}
    \textbf{LIBERO} & & \textbf{Spatial} & \textbf{Object} & \textbf{Goal} & \textbf{Long} & \multicolumn{2}{c}{\textbf{Avg.}} \\
    \midrule
    & Qwen2.5-VL-GR00T & 94.6 & 97.2 & 93.8 & 90.2 & \multicolumn{2}{c}{93.9} \\
    \rowcolor{blue!4}
    & \tb{Qwen2.5-VL-GR00T+TokenPB} & 96.6 & \ul{98.8} & 97.0 & 94.6 & \multicolumn{2}{c}{96.8} \\
    & InternVL3.5-GR00T & 95.8 & 96.0 & 94.2 & 92.8 & \multicolumn{2}{c}{94.7} \\
    \rowcolor{blue!4}
    & \tb{InternVL3.5-GR00T+TokenPB} & 97.6 & \tb{99.4} & \ul{97.2} & \ul{96.8} & \multicolumn{2}{c}{\ul{97.8}} \\
    & $\pi_{0.5}$ & \ul{97.8} & 98.4 & 96.6 & 93.0 & \multicolumn{2}{c}{96.5} \\
    \rowcolor{blue!4}
    & \tb{$\pi_{0.5}$+TokenPB} & \tb{98.4} & \tb{99.4} & \tb{98.2} & \tb{97.6} & \multicolumn{2}{c}{\tb{98.4}} \\
    \midrule
    \rowcolor{gray!8}
    \textbf{CALVIN ABC$\to$D} & & \textbf{Task 1} & \textbf{Task 2} & \textbf{Task 3} & \textbf{Task 4} & \textbf{Task 5} & \textbf{Avg.\ Len.} \\
    \midrule
    & Qwen2.5-VL-GR00T & 90.8 & 81.2 & 74.9 & 68.7 & 59.5 & 3.751 \\
    \rowcolor{blue!4}
    & \tb{Qwen2.5-VL-GR00T+TokenPB} & 93.1 & \ul{87.4} & \ul{81.6} & \ul{75.9} & \ul{71.3} & \ul{4.093} \\
    & $\pi_{0.5}$ & \ul{93.5} & 86.7 & 79.6 & 75.3 & 71.1 & 4.071 \\
    \rowcolor{blue!4}
    & \tb{$\pi_{0.5}$+TokenPB} & \tb{95.4} & \tb{90.6} & \tb{85.3} & \tb{79.6} & \tb{75.8} & \tb{4.267} \\
    \midrule
    \rowcolor{gray!8}
    \textbf{RoboTwin 2.0} & & \textbf{Block Rank RGB} & \textbf{Open Laptop} & \textbf{Adjust Bottle} & \textbf{......} & \multicolumn{2}{c}{\textbf{Avg.}} \\
    & & {\scriptsize\textit{Easy\,/\,Hard}} & {\scriptsize\textit{Easy\,/\,Hard}} & {\scriptsize\textit{Easy\,/\,Hard}} & & \multicolumn{2}{c}{\scriptsize\textit{Easy\,/\,Hard}} \\
    \midrule
    & Qwen2.5-VL-GR00T & 38.0\,/\,11.0 & 39.0\,/\,18.0 & 96.0\,/\,16.0 & \ldots & \multicolumn{2}{c}{30.1\,/\,\phantom{0}6.7} \\
    \rowcolor{blue!4}
    & \tb{Qwen2.5-VL-GR00T+TokenPB} & \tb{52.0}\,/\,\tb{24.0} & \tb{55.0}\,/\,\tb{27.0} & \tb{97.0}\,/\,\tb{26.0} & \ldots & \multicolumn{2}{c}{\tb{39.3}\,/\,\tb{10.2}} \\
    \midrule
    \rowcolor{gray!8}
    \textbf{Real-World} & & \textbf{Push Blocks} & \textbf{Clean Table} & \textbf{Pick \& Place} & \textbf{Hang Cups} & \multicolumn{2}{c}{\textbf{Avg.}} \\
    & & {\scriptsize\textit{Seen\,/\,Unseen}} & {\scriptsize\textit{Seen\,/\,Unseen}} & {\scriptsize\textit{Seen\,/\,Unseen}} & {\scriptsize\textit{Seen\,/\,Unseen}} & \multicolumn{2}{c}{\scriptsize\textit{Seen\,/\,Unseen}} \\
    \midrule
    & Qwen2.5-VL-GR00T & 56.2\,/\,\phantom{0}5.0 & 52.5\,/\,\phantom{0}8.0 & 45.5\,/\,28.0 & 43.7\,/\,32.0 & \multicolumn{2}{c}{49.5\,/\,18.3} \\
    \rowcolor{blue!4}
    & \tb{Qwen2.5-VL-GR00T+TokenPB} & \ul{75.4}\,/\,\tb{45.0} & \tb{72.5}\,/\,\ul{24.0} & 75.0\,/\,\ul{36.0} & 62.5\,/\,40.0 & \multicolumn{2}{c}{71.4\,/\,\ul{36.3}} \\
    & InternVL3.5-GR00T & 54.1\,/\,10.0 & 56.7\,/\,\phantom{0}8.0 & 55.0\,/\,16.0 & 68.7\,/\,40.0 & \multicolumn{2}{c}{58.6\,/\,18.5} \\
    \rowcolor{blue!4}
    & \tb{InternVL3.5-GR00T+TokenPB} & 72.7\,/\,\ul{40.0} & 66.7\,/\,16.0 & \tb{83.3}\,/\,32.0 & \tb{81.3}\,/\,\tb{56.0} & \multicolumn{2}{c}{\ul{76.1}\,/\,36.0} \\
    & $\pi_{0.5}$ & 74.0\,/\,25.0 & 61.5\,/\,16.0 & 71.3\,/\,\tb{40.0} & 55.3\,/\,40.0 & \multicolumn{2}{c}{65.6\,/\,30.3} \\
    \rowcolor{blue!4}
    & \tb{$\pi_{0.5}$+TokenPB} & \tb{81.7}\,/\,\ul{40.0} & \ul{70.7}\,/\,\tb{48.0} & \ul{79.5}\,/\,\tb{40.0} & \ul{78.3}\,/\,\ul{48.0} & \multicolumn{2}{c}{\tb{77.6}\,/\,\tb{44.0}} \\
    \bottomrule
    \end{tabular}
    }
    \end{table*}
    
We evaluate TokenPB across simulated and real-world manipulation benchmarks, addressing four research questions that map one-to-one to the subsections below:

\textbf{Q1:} Does TokenPB consistently improve over single-sample inference across benchmarks, backbones, and distribution shifts? (Section~\ref{sec:main_result})

\textbf{Q2:} How does TokenPB compare to external verification methods in both task performance and computational cost? (Section~\ref{sec:external_comparison})

\textbf{Q3:} How does the verification overhead scale with the candidate set size $M$? (Section~\ref{sec:cost_tradeoff})

\textbf{Q4:} Where do the gains come from---co-training, test-time verification, or both---and can the bucket score reliably distinguish good actions from bad ones? (Section~\ref{sec:analysis})

\vpara{Setup.}
We evaluate two VLM backbones (Qwen2.5-VL-3B, InternVL3.5-2B) paired with a GR00T-N1-style
flow-matching action expert~\citep{bjorck2025gr00t}, and one VLA ($\pi_{0.5}$) following the same
co-training protocol. All models are co-trained in a single checkpoint with a joint objective combining
(i) the flow-matching loss $\mathcal{L}_{\mathrm{fm}}$ for continuous action generation and
(ii) an auxiliary bucket-token objective $\mathcal{L}_{\text{bucket}}$ for candidate scoring.
At inference time, \textsc{Model-GR00T} denotes the resulting backbone+expert policy run with
single-sample inference ($M{=}1$) and no selection, while \textsc{Model-GR00T+TokenPB} enables
test-time sampling-and-selection by scoring $M$ candidates (default $M{=}16$) using the learned bucket
score. Simulation benchmarks include LIBERO~\cite{liu2023libero}, CALVIN~\cite{mees2022calvin}, and
RoboTwin2.0~\cite{chen2025robotwin}; real-world evaluation covers four bimanual tasks
(Appendix~\ref{metric},~\ref{sec:realworldtasks}).


\subsection{Comparison with Single-Sample vs.\ Test-Time Inference}
\label{sec:main_result}

Table~\ref{table:single_vs_tti} tests the central premise of TokenPB: from the \emph{same co-trained
checkpoint}, switching from single-sample inference ($M{=}1$) to test-time sampling-and-selection
(with TokenPB scoring, default $M{=}16$) should consistently improve task success without any
additional training. This is confirmed across every benchmark--backbone combination. On LIBERO the average gain is +2.7 points (up to +3.1 on InternVL3.5); on CALVIN ABC$\to$D, +0.34 in average  length on Qwen2.5-VL and +0.20 on $\pi_{0.5}$; on RoboTwin2.0, +9.2 on Easy and +3.5 on Hard; and in real-world environments, +21.9 / +17.5 / +12.0 on the three backbones (average +17.1). The gains also generalize beyond the training distribution: on a held-out Real-World Unseen setting with novel instructions and object configurations, TokenPB yields an average +16.4-point gain---comparable to the in-distribution improvement---indicating that the verifier captures task-relevant action semantics rather than memorizing training-time patterns (statistical details in Appendix~\ref{sec:realworld_stats}).

\subsection{Comparison with Baseline Verification Methods}
\label{sec:external_comparison}


A central claim of TokenPB is that it enables effective test-time sampling-and-selection \emph{without}
relying on a separately trained external verifier, while achieving competitive or superior task
performance. We validate this claim along two axes: \emph{task performance} and
\emph{computational cost}, comparing against both external-verifier-based selection (e.g., RoboMonkey)
and internal scoring baselines that still incur per-candidate scoring compute (e.g., MG-Select).

\vpara{Task performance.} We compare TokenPB against RoboMonkey~\cite{robomonkey2025}, MG-Select~\cite{mgselect2025}, AsyncVLA \cite{asyncvla2025} and RoVer~\cite{rover2025} on their respective benchmarks. TokenPB outperforms MG-Select on LIBERO (96.8\% vs.\ 93.1\%) and achieves the highest success rate on SimplerEnv-WidowX (0.76), as shown in Table~\ref{table:compare_libero_simplerenv}. It also yields a larger gain than RoVer on CALVIN (+0.20 vs.\ +0.11 average length, Table~\ref{table:compare_calvin})---notably, TokenPB achieves these results without requiring any separately trained verifier.

\begin{wrapfigure}{r}{0.4\textwidth}
  \centering
  \includegraphics[width=0.38\textwidth]{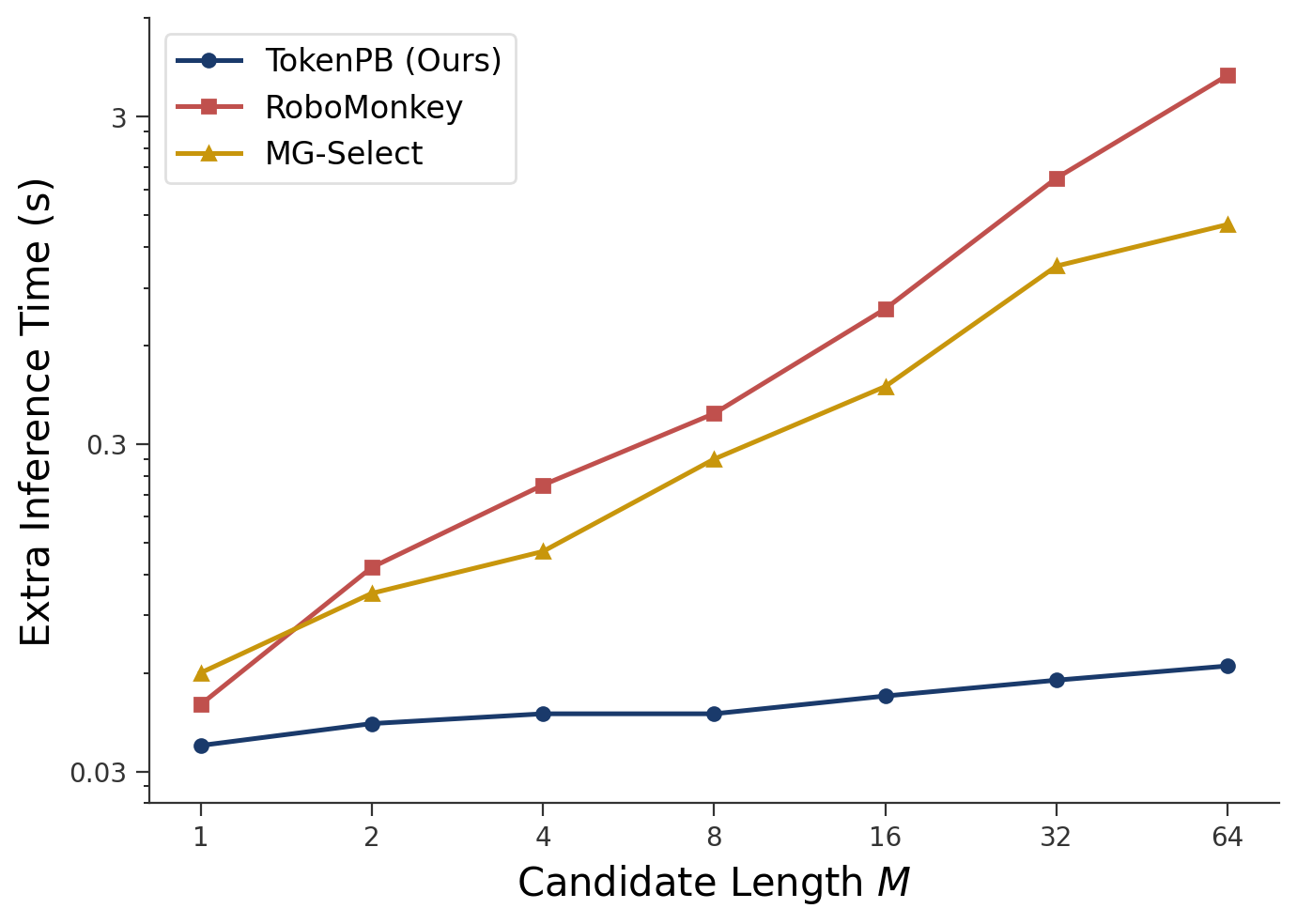}
  \caption{\textbf{Inference time comparison with RoboMonkey and MG-Select.}}
  \label{fig:inference_time}
\end{wrapfigure}


\vpara{Computational cost.}
Figure~\ref{fig:inference_time} compares the extra inference time as $M$ varies from 1 to 64.
RoboMonkey and MG-Select grow roughly linearly with $M$ ($>$4\,s and $>$1.5\,s at $M{=}64$), because
their scoring introduces \emph{per-candidate} compute: RoboMonkey runs a separately trained verifier on
each candidate action, while MG-Select performs an additional \emph{masked} scoring forward pass for
each candidate under the VLA's own masking distribution.
In both cases, batching can improve throughput but does not eliminate the $O(M)$ \emph{total} scoring
compute, since each candidate still requires its own scoring computation (and MG-Select cannot score all
candidates in one teacher-forced continuation pass conditioned on a shared observation-prefix KV cache).
In contrast, TokenPB stays below 0.2\,s even at $M{=}64$---roughly one order of magnitude faster than
RoboMonkey and MG-Select. The key reason is that TokenPB's sampling and verification share the same observation prefix
within a single VLM backbone: the observation-prefix KV cache can be computed once and reused to score
all $M$ candidates via a single batched teacher-forced pass over candidate-specific token continuations,
yielding near-constant overhead with respect to $M$ and moderate memory growth due to KV-cache reuse
(Figure~\ref{fig:cost_tradeoff}(left) in Section~\ref{sec:cost_tradeoff}).

The timing comparison enables the respective prefix-sharing and batching optimizations; additional protocol details and Qwen2.5-VL-3B memory measurements are provided in Appendix~\ref{app:efficiency}.

\begin{table}[t]
    \caption{\textbf{Comparison with baseline verification methods on LIBERO and SimplerEnv-WidowX.} Original baseline rows retain results from their official publications; only LIBERO-Long is available for the original RoboMonkey row, and V-GPS is goal-conditioned. Added rows report MG-Select single-sample inference and a reproduced RoboMonkey-Qwen2.5 verifier with the shared $\pi_{0.5}$ single-sample baseline. Reproduction details are in Appendix~\ref{app:baseline_protocol}.}
    \centering
    \resizebox{\linewidth}{!}{
    \begin{tabular}{l | c c c c c || c c c c c}
    \toprule
    & \multicolumn{5}{c||}{\textbf{LIBERO (\%)}} & \multicolumn{5}{c}{\textbf{SimplerEnv-WidowX (SR)}} \\
    \textbf{Method} & \textbf{Spatial} & \textbf{Object} & \textbf{Goal} & \textbf{Long} & \textbf{Avg.} & \textbf{Spoon on Towel} & \textbf{Carrot on Plate} & \textbf{Stack Cubes} & \textbf{Eggplant in Basket} & \textbf{Avg.} \\
    \midrule
    V-GPS & -- & -- & -- & -- & -- & 0.46 & 0.16 & 0.07 & 0.84 & 0.38 \\
    RoboMonkey & -- & -- & -- & 56.5 & -- & 0.68 & 0.72 & 0.44 & 0.18 & 0.51 \\
    MG-Select (single) & 97.4 & 95.4 & 95.6 & 79.6 & 92.0 & 0.67 & 0.71 & 0.42 & 0.08 & 0.47 \\
    MG-Select & 97.2 & 98.0 & 94.5 & 82.7 & 93.1 & 0.69 & \tb{0.75} & 0.43 & 0.14 & 0.50 \\
    AsyncVLA & 98.0 & 98.2 & 96.4 & 96.0 & 97.4 & 0.71 & 0.67 & \tb{0.58} & 0.88 & 0.71 \\
    \midrule
    $\pi_{0.5}$ (single, shared) & 97.8 & 98.4 & 96.6 & 93.0 & 96.5 & 0.74 & 0.65 & 0.38 & 0.87 & 0.66 \\
    RoboMonkey-Qwen2.5 & \tb{98.8} & 98.2 & 97.0 & 95.0 & 97.3 & 0.76 & 0.65 & 0.49 & 0.95 & 0.71 \\
    \rowcolor{blue!4}
    \tb{TokenPB (Ours)} & \tb{98.4} & \tb{99.4} & \tb{98.2} & \tb{97.6} & \tb{98.4} & \tb{0.86} & 0.71 & 0.47 & \tb{1.00} & \tb{0.76} \\
    \bottomrule
    \end{tabular}
    }
    \label{table:compare_libero_simplerenv}
    \end{table}
    
    \begin{table}[t]
    \caption{\textbf{Comparison with baseline verification methods on CALVIN ABC$\to$D.} All baseline results are from their respective official publications.}
    \centering
    \resizebox{0.85\linewidth}{!}{
    \begin{tabular}{l | c c c c c c}
    \toprule
    \textbf{Method} & \textbf{Task 1} & \textbf{Task 2} & \textbf{Task 3} & \textbf{Task 4} & \textbf{Task 5} & \textbf{Avg. Len.} \\
    \midrule
    MoDE & 96.2 & 88.9 & 81.1 & 71.8 & 63.5 & 4.01 \\
    MoDE + RoVer & 97.1 & 90.9 & 82.5 & 74.9 & 66.6 & 4.12 {\scriptsize(+0.11)} \\
    \midrule
    $\pi_{0.5}$ & 93.5 & 86.7 & 79.6 & 75.3 & 71.1 & 4.07 \\
    \rowcolor{blue!4}
    \tb{$\pi_{0.5}$ + TokenPB (Ours)} & \tb{95.4} & \tb{90.6} & \tb{85.3} & \tb{79.6} & \tb{75.8} & \tb{4.27} {\scriptsize\tb{(+0.20)}} \\
    \bottomrule
    \end{tabular}
    }
    \label{table:compare_calvin}
    \end{table}

\subsection{Test-Time Scaling Behavior}
\label{sec:cost_tradeoff}
\begin{figure}[t]
  \centering
  \includegraphics[width=\textwidth]{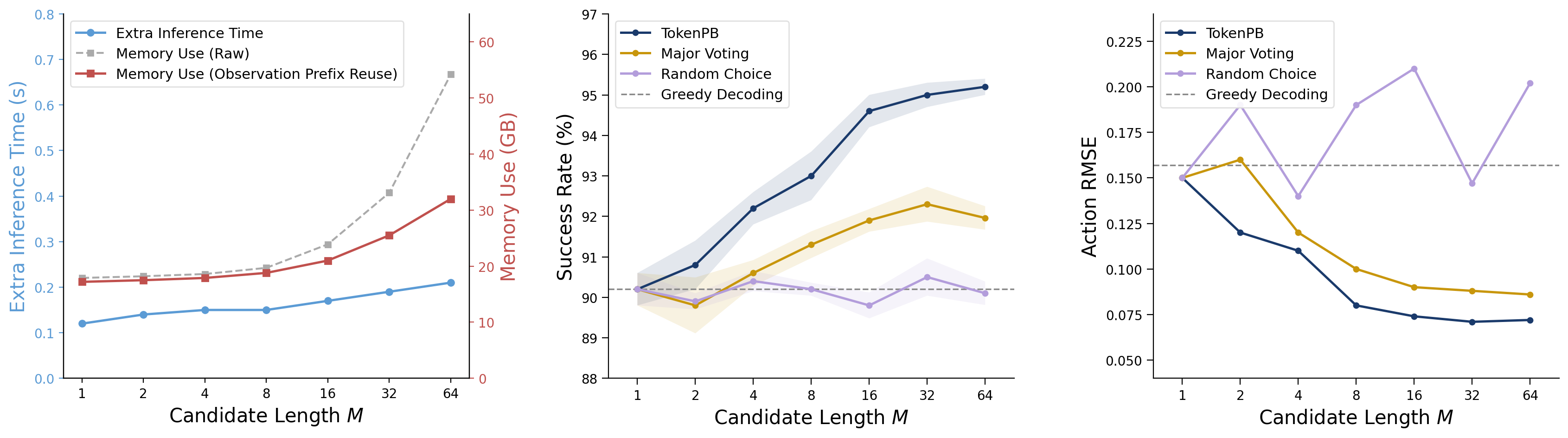}
  \caption{\textbf{Test-time cost--performance trade-off on LIBERO-Long as $M$ varies from 1 to 64.} \emph{Left:} Extra inference time (blue, left axis) remains nearly constant across all $M$. The gray dashed line shows raw GPU memory without optimization; by reusing the observation-prefix KV cache (red), memory grows only to $\sim$32\,GB even at $M{=}64$---roughly half the raw cost. \emph{Middle:} Success rate versus $M$. TokenPB (blue) improves steadily, whereas random choice (purple) and majority voting (yellow) plateau. \emph{Right:} Action RMSE versus $M$. TokenPB drives a monotonic decrease, confirming that selected actions become progressively closer to the ground truth. We adopt $M{=}16$ as the default.}
  \label{fig:cost_tradeoff}
\end{figure}

We study how TokenPB's performance and overhead scale with $M$ on LIBERO-Long (Figure~\ref{fig:cost_tradeoff}). Because all candidates share the same observation prefix and scoring is batched in a single forward pass, the extra inference time remains nearly constant regardless of $M$. A naïve implementation would duplicate the full KV cache for each candidate, causing raw memory to exceed 60\,GB at $M{=}64$. By reusing the observation-prefix KV cache---which is identical across all candidates---we reduce memory to ${\sim}$32\,GB at $M{=}64$ and ${\sim}$20\,GB at $M{=}16$, comfortably fitting on a single consumer-grade GPU. We adopt $M{=}16$ as the default.

Performance improves steadily with $M$: success rate climbs from ${\sim}$90\% ($M{=}1$) to over 95\% ($M{=}32$), while action RMSE decreases monotonically. In contrast, random choice and majority voting plateau or fluctuate and fail to benefit from larger candidate pools. TokenPB thus converts additional test-time candidates into tangible gains at nearly fixed inference cost.

\subsection{Ablation and Analysis}
\label{sec:analysis}

\begin{table*}[t]
\centering
\caption{\textbf{Disentangling co-training representation benefit from test-time verification benefit.} All variants use Qwen2.5-VL-3B as the VLM backbone. ``Co-trained + Single ($M{=}1$)'' corresponds to the GR00T baseline in Table~\ref{table:single_vs_tti}. ``No Bucket'' removes the auxiliary token-bucket cross-entropy $\mathcal{L}_{\text{bucket}}$ during training
(the action expert is still trained with $\mathcal{L}_{\text{fm}}$); consequently, no learned bucket-likelihood
scorer is available at test time and multi-sample selection reduces to random choice. ``Full TokenPB'' is the complete framework with co-training and sampling-and-selection ($M{=}16$).}
\label{tab:ablation_disentangle}
\renewcommand{\arraystretch}{1.2}
\resizebox{\linewidth}{!}{
\begin{tabular}{@{}l  c c c c c  c c c c c c@{}}
\toprule
\multirow{3}{*}{\textbf{Configuration}} & \multicolumn{5}{c}{\textbf{LIBERO (\%)}} & \multicolumn{6}{c}{\textbf{RoboTwin2.0 (\%)}} \\
\cmidrule(lr){2-6} \cmidrule(lr){7-12}
& \multirow{2}{*}{\textbf{Spatial}} & \multirow{2}{*}{\textbf{Object}} & \multirow{2}{*}{\textbf{Goal}} & \multirow{2}{*}{\textbf{Long}} & \multirow{2}{*}{\textbf{Avg.}} & \multicolumn{2}{c}{\textbf{Blocks Ranking}} & \multicolumn{2}{c}{\textbf{Open Laptop}} & \multicolumn{2}{c}{\textbf{Shake Bottle}} \\
\cmidrule(lr){7-8} \cmidrule(lr){9-10} \cmidrule(lr){11-12}
& & & & & & {\small Easy} & {\small Hard} & {\small Easy} & {\small Hard} & {\small Easy} & {\small Hard} \\
\midrule
No Bucket + Single ($M{=}1$) & 92.4 & 95.6 & 91.8 & 87.6 & 91.9 & 32 & 7 & 33 & 13 & 79 & 16 \\
No Bucket + Random Select ($M{=}16$) & 92.0 & 95.2 & 91.4 & 87.8 & 91.6 & 30 & 6 & 31 & 12 & 77 & 15 \\
\midrule
Co-trained + Single ($M{=}1$) & 93.6 & 96.2 & 92.8 & 90.2 & 93.2 & 38 & 11 & 39 & 18 & 81 & 21 \\
\midrule
\rowcolor{blue!4}
\tb{Full TokenPB ($M{=}16$)} & \tb{96.6} & \tb{98.8} & \tb{97.0} & \tb{94.6} & \tb{96.8} & \tb{52} & \tb{24} & \tb{55} & \tb{27} & \tb{86} & \tb{37} \\
\bottomrule
\end{tabular}
}
\end{table*}

\vpara{Disentangling co-training and verification.} We isolate the two components by removing each in turn (Table~\ref{tab:ablation_disentangle}). Without the bucket objective, randomly selecting among candidates provides no gain over a single sample---candidate diversity alone is useless without a scoring function. Adding the bucket objective during training already improves success rate by +2.0 on LIBERO and +4--6 on RoboTwin2.0. Applying verification on top yields a further +2.9 on LIBERO and up to +16 on RoboTwin2.0 Hard. The total gain exceeds the sum of individual components, confirming that co-training provides the scoring function and verification exploits it at test time.

\begin{table*}[t]
    \centering
    \caption{\textbf{Fine-grained trajectory metrics on RoboTwin2.0 (Hard).} TokenPB-selected actions are compared against single-sample ones from the same co-trained checkpoint. \emph{Recovery Attempts}: higher values indicate more retry attempts when the initial action fails, reflecting stronger error-correction behavior. \emph{Env.\ Collision Before Grasp}: lower values indicate fewer irrelevant collisions with the environment before grasping the target object.}
    \label{table:diffusion_1}
    \renewcommand{\arraystretch}{1.25}
    \resizebox{\linewidth}{!}{
    \begin{tabular}{@{}l  cc  cc  cc  cc@{}}
    \toprule
    \multirow{2}{*}{\textbf{Metric}} & \multicolumn{2}{c}{\textbf{Adjust Bottle}} & \multicolumn{2}{c}{\textbf{Blocks Ranking RGB}} & \multicolumn{2}{c}{\textbf{Open Laptop}} & \multicolumn{2}{c}{\textbf{Shake Bottle}} \\
    \cmidrule(lr){2-3} \cmidrule(lr){4-5} \cmidrule(lr){6-7} \cmidrule(lr){8-9}
    & \textbf{TokenPB} & \textbf{Baseline} & \textbf{TokenPB} & \textbf{Baseline} & \textbf{TokenPB} & \textbf{Baseline} & \textbf{TokenPB} & \textbf{Baseline} \\
    \midrule
    Avg.\ Success Steps ($\downarrow$)
    & \tb{107.3}~{\scriptsize(-45.2)} & 152.5
    & \tb{754.0}~{\scriptsize(-282.9)} & 1036.9
    & \tb{142.3}~{\scriptsize(-17.1)} & 159.4
    & \tb{95.3}~{\scriptsize(-54.4)} & 149.7 \\
    \addlinespace[4pt]
    Recovery Attempts ($\uparrow$)
    & \tb{1.1}~{\scriptsize(+0.7)} & 0.4
    & \tb{3.3}~{\scriptsize(+2.6)} & 0.7
    & \tb{2.9}~{\scriptsize(+2.6)} & 0.3
    & \tb{1.3}~{\scriptsize(+0.6)} & 0.7 \\
    \addlinespace[4pt]
    Env.\ Collision Before Grasp (\%,\,$\downarrow$)
    & \tb{3}~{\scriptsize(-43)} & 46
    & \tb{14}~{\scriptsize(-15)} & 29
    & \tb{0}~{\scriptsize(-14)} & 14
    & \tb{4}~{\scriptsize(-34)} & 38 \\
    \midrule
    Success Rate (\%,\,$\uparrow$)
    & \tb{26}~{\scriptsize(+10)} & 16
    & \tb{24}~{\scriptsize(+13)} & 11
    & \tb{27}~{\scriptsize(+9)} & 18
    & \tb{37}~{\scriptsize(+16)} & 21 \\
    \bottomrule
    \end{tabular}
    }
    \end{table*}
    
\begin{figure}[t]
  \centering
  \includegraphics[width=\textwidth]{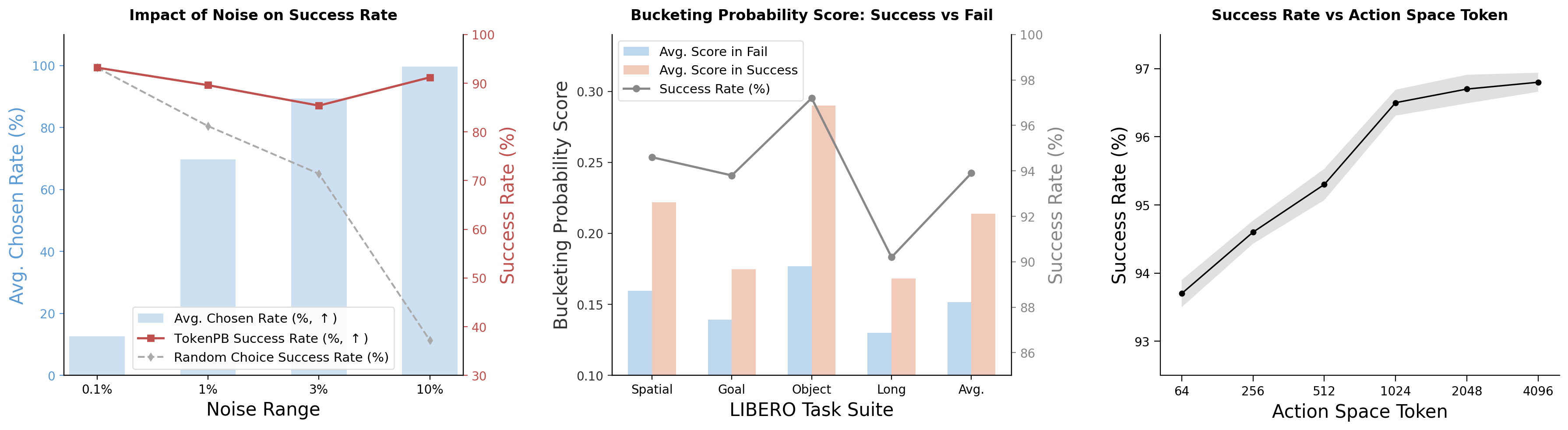}
  \caption{\textbf{Verification effectiveness analysis.} Left: TokenPB maintains near-constant success rate under noise injection compared to random choice. Middle: Bucketing probability scores of successful trajectories are significantly higher than failed ones, and the score magnitude tracks task success rate across suites. Right: Task success rate saturates beyond 1024 buckets.}
  \label{fig:table_charts}
\end{figure}

\vpara{Score discriminability.} We validate the bucket score from two angles (Figure~\ref{fig:table_charts}). First, injecting random noise into all but one clean candidate on LIBERO (left panel), the verifier's clean-candidate selection rate rises to nearly 100\% at 10\% noise magnitude while random choice degrades sharply. Second, successful trajectories consistently exhibit significantly higher scores than failed ones across all LIBERO suites (middle panel); moreover, the relative ordering of average scores across suites mirrors the ordering of task success rates, confirming that the score reliably reflects task difficulty. TokenPB-selected actions on RoboTwin2.0 Hard further improve success rate by +12.0 points with fewer collisions and more recovery attempts (Table~\ref{table:diffusion_1}). Qualitative examples are in Appendix~\ref{app:qualitative}; additional backbone, LIBERO-Plus, RoboCasa, and embodiment-pretraining evaluations are in Appendix~\ref{app:backbones}.

\vpara{Bucket granularity.} We vary the vocabulary size from 64 to 4096 to examine how discretization resolution affects verification performance (Figure~\ref{fig:table_charts}, right). Success rate improves monotonically and saturates beyond 1024 tokens at approximately 97\%. With a small vocabulary, coarse discretization can map distinct continuous actions to the same token sequence, limiting the verifier's ability to distinguish candidates with different execution quality. Increasing the vocabulary provides finer action representations and allows the bucket score to capture distinctions that would otherwise be lost during tokenization. However, further increasing the vocabulary beyond 1024 yields only marginal gains, suggesting that finer discretization alone is insufficient to substantially improve selection. This trend is consistent with discretization becoming a less restrictive bottleneck at sufficient granularity, with further improvements depending more on the quality of the sampled candidates and the verifier's ability to rank them accurately.

\section{Limitations and Discussion}
\label{app:limitations}

\vpara{Memory scaling with candidate set size.} Although observation-prefix KV-cache reuse substantially reduces per-candidate cost, GPU memory still grows linearly with the number of candidates $M$, which may limit deployment on memory-constrained edge devices.

\vpara{Frequency-based scoring and multi-modal action distributions.} When multiple valid actions exist for the same observation, the verifier favors the most \emph{frequent} demonstration action rather than the globally optimal one, since the scoring function is trained to approximate the conditional demonstration distribution $P^*(\cdot \mid s)$. However, this concern is mitigated by the observation-conditioned nature of the bucket distribution: in states where atypical actions (e.g., obstacle avoidance detours) are appropriate, these actions are well-represented in the \emph{conditional} distribution $P^*(c \mid s_t)$ for the relevant observations, so the verifier does not penalize them. Empirically, the Real-World Unseen gains (+16.4 points, Table~\ref{table:single_vs_tti}) are comparable to in-distribution gains (+17.1), and the increased recovery attempts on RoboTwin2.0 Hard (Table~\ref{table:diffusion_1}) confirm that the verifier successfully identifies non-typical corrective actions rather than blindly favoring high-frequency patterns.

\vpara{Dependence on demonstration data coverage.} For action types entirely absent from the training demonstrations, TokenPB cannot produce meaningful scores---but this is a fundamental data coverage limitation shared by all behavior-cloning-based VLA methods.

\section{Conclusion}
We propose Token Probability Bucketing (TokenPB), a self-verification framework that unifies action generation and verification within a single VLA checkpoint. By co-training the VLM backbone to predict observation-conditioned bucket distributions, TokenPB enables test-time candidate selection without any external verifier, achieving consistent gains across multiple architectures and benchmarks with near-constant overhead.

\bibliographystyle{plainnat}
\bibliography{custom}
\newpage
\appendix

\section{Qualitative Verification Examples}
\label{app:qualitative}

\begin{figure}[t]
  \centering
  \includegraphics[width=\textwidth]{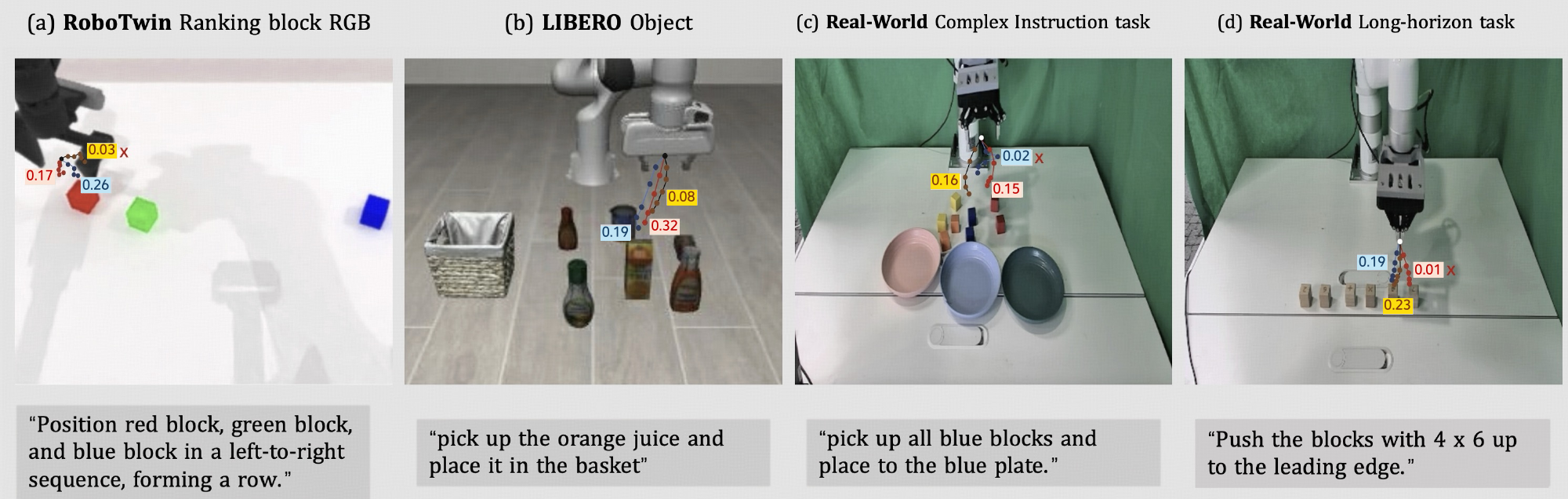}
  \caption{\textbf{Qualitative visualization of TokenPB verification across diverse tasks.} Each panel shows a task scene with candidate actions and their bucketing probability scores. Candidates with higher scores (shown in blue boxes with check marks) are selected for execution, while lower-scored candidates (red boxes with crosses) are rejected. TokenPB correctly assigns the highest probability to semantically appropriate actions: (a)~placing blocks in the correct spatial order in RoboTwin, (b)~grasping the target object in LIBERO, (c)~selecting the correct objects in a complex real-world instruction task, and (d)~pushing the correct blocks in a long-horizon task.}
  \label{fig:qualitative}
\end{figure}

To provide an intuitive understanding of the verification mechanism, Figure~\ref{fig:qualitative} visualizes representative examples of TokenPB scoring across diverse tasks from RoboTwin, LIBERO, and real-world environments. In each case, the verifier assigns the highest bucketing probability score to the candidate whose action semantics best match the task instruction, while candidates with incorrect spatial targets or object selections receive substantially lower scores. These examples illustrate that the co-trained VLM backbone captures task-relevant action semantics through the bucket probability distribution, enabling reliable best-candidate selection even in visually complex multi-object scenes.

\section{Scoring Monotonicity under Cross-Entropy Training}
\label{app:scoring_theory}

We establish a stronger result: the cross-entropy training dynamics guarantee that the predicted probabilities eventually become monotone in the true probabilities, even before exact calibration is achieved.

\subsection{Finite-Sample Convergence}

\begin{assumption}[Realizability]
\label{assump:realizability}
The model class $\{\pi_\theta : \theta \in \Theta\}$ contains the true conditional distribution, i.e., there exists $\theta^* \in \Theta$ such that $\pi_{\theta^*}(\cdot \mid s) = P^*(\cdot \mid s)$ for all $s$ in the support of the training distribution.
\end{assumption}

\begin{assumption}[Regularity]
\label{assump:regularity}
The loss landscape $\mathcal{L}_{\text{CE}}(\theta)$ satisfies: (i) $\nabla^2 \mathcal{L}_{\text{CE}}(\theta^*)$ is positive definite (non-degenerate minimum); (ii) $\pi_\theta(c \mid s)$ is continuous in $\theta$ for all $c, s$.
\end{assumption}

\begin{theorem}[Scoring monotonicity]
\label{thm:scoring_monotonicity}
Under Assumptions~\ref{assump:realizability} and~\ref{assump:regularity}, suppose that gradient-based optimization on $\mathcal{L}_{\text{CE}}$ converges to a neighborhood of $\theta^*$, i.e., $\|\theta_T - \theta^*\| \leq \epsilon$ for sufficiently large $T$. Then for any observation $s$ and any two candidate actions $a_m, a_{m'}$ with $P^*(h(a_m) \mid s) > P^*(h(a_{m'}) \mid s)$ (i.e., the sequence-level probability of $a_m$'s token sequence is higher under the true distribution), there exists $\epsilon_0 > 0$ such that for all $\epsilon < \epsilon_0$:
\begin{equation}
    \log \pi_{\theta_T}\bigl(h(a_m) \mid s\bigr) > \log \pi_{\theta_T}\bigl(h(a_{m'}) \mid s\bigr).
\end{equation}
That is, the verifier's log-probability score preserves the true quality ranking at the sequence level.
\end{theorem}

\begin{proof}
Recall that the sequence-level probability factorizes as $\pi_\theta(h(a)\mid s) = \prod_{\ell=1}^L \pi_\theta(c_\ell \mid s, c_{<\ell})$, hence $\log\pi_\theta(h(a)\mid s) = \sum_{\ell=1}^L \log\pi_\theta(c_\ell\mid s,c_{<\ell})$. By Assumption~\ref{assump:realizability}, $\pi_{\theta^*}(h(a)\mid s) = P^*(h(a)\mid s)$ for all $a,s$.

By continuity of $\pi_\theta(c_\ell \mid s, c_{<\ell})$ in $\theta$ (Assumption~\ref{assump:regularity}(ii)) and the composition with the continuous $\log$ function (noting that probabilities are bounded away from zero on the support of $P^*$), $\log\pi_\theta(h(a)\mid s)$ is continuous in $\theta$ for any fixed $a,s$. Therefore, for any $\delta > 0$ there exists $\epsilon_0 > 0$ such that $\|\theta - \theta^*\| < \epsilon_0$ implies $|\log\pi_\theta(h(a)\mid s) - \log\pi_{\theta^*}(h(a)\mid s)| < \delta$ for all $a, s$ in the support.

Since $\pi_{\theta^*}(h(a_m)\mid s) = P^*(h(a_m)\mid s) > P^*(h(a_{m'})\mid s) = \pi_{\theta^*}(h(a_{m'})\mid s)$, we have $\log P^*(h(a_m)\mid s) > \log P^*(h(a_{m'})\mid s)$. Let $\gamma = \log P^*(h(a_m)\mid s) - \log P^*(h(a_{m'})\mid s) > 0$. Choosing $\delta = \gamma/3$ and the corresponding $\epsilon_0$, for $\|\theta_T - \theta^*\| < \epsilon_0$:
\begin{align}
    \log\pi_{\theta_T}(h(a_m) \mid s) &> \log P^*(h(a_m) \mid s) - \gamma/3, \\
    \log\pi_{\theta_T}(h(a_{m'}) \mid s) &< \log P^*(h(a_{m'}) \mid s) + \gamma/3.
\end{align}
Therefore $\log\pi_{\theta_T}(h(a_m) \mid s) - \log\pi_{\theta_T}(h(a_{m'}) \mid s) > \gamma - 2\gamma/3 = \gamma/3 > 0$.
\end{proof}

\subsection{Relaxation of Realizability}

In practice, the model class may not contain the exact Bayes predictor. We show that the result holds approximately under misspecification.

\begin{proposition}[Approximate monotonicity under misspecification]
\label{prop:misspecification}
Without Assumption~\ref{assump:realizability}, let $\theta^* = \arg\min_{\theta \in \Theta} \mathcal{L}_{\text{CE}}(\theta)$ be the best-in-class minimizer. Define the log-space approximation error $\varepsilon_{\mathrm{approx}} = \max_{s,a} |\log\pi_{\theta^*}(h(a) \mid s) - \log P^*(h(a) \mid s)|$ and the optimization residual $\epsilon_{\mathrm{opt}} = \max_{s,a}|\log\pi_{\theta_T}(h(a) \mid s) - \log\pi_{\theta^*}(h(a) \mid s)|$. Then the scoring monotonicity holds for all candidate pairs satisfying $\log P^*(h(a_m) \mid s) - \log P^*(h(a_{m'}) \mid s) > 2(\varepsilon_{\mathrm{approx}} + \epsilon_{\mathrm{opt}})$.
\end{proposition}

\begin{proof}
By triangle inequality in log-space:
\begin{align}
    &|\log\pi_{\theta_T}(h(a) \mid s) - \log P^*(h(a) \mid s)| \notag\\
    &\quad\leq |\log\pi_{\theta_T}(h(a) \mid s) - \log\pi_{\theta^*}(h(a) \mid s)| \notag+ |\log\pi_{\theta^*}(h(a) \mid s) - \log P^*(h(a) \mid s)| \\
    &\quad\leq \epsilon_{\mathrm{opt}} + \varepsilon_{\mathrm{approx}}.
\end{align}
Applying the same argument as Theorem~\ref{thm:verifier_soundness} with total log-space error $\epsilon_{\mathrm{opt}} + \varepsilon_{\mathrm{approx}}$ replacing $\varepsilon_{\mathrm{cal}}$ yields the result.
\end{proof}

\begin{remark}[Approximation and optimization]
Proposition~\ref{prop:misspecification} separates limitations of the model class from incomplete optimization. Increasing model capacity may reduce approximation error, whereas improving optimization may reduce the residual relative to the best-in-class predictor. Ranking is preserved only when the candidate log-probability gap exceeds twice their combined error.
\end{remark}

\section{Proof of Verifier Soundness via Calibration Theory}
\label{app:calibration_proof}

We establish that the cross-entropy-trained bucket head provides a theoretically grounded scoring function for action verification. The analysis is architecture-agnostic and applies directly to the VLMs used in our experiments; the guarantees are conditional on the stated assumptions and on sufficiently accurate estimation of the demonstration distribution.

\subsection{Setup and Notation}

Let the training set be $\mathcal{D} = \{(s_j, a_j)\}_{j=1}^n$, where $s_j$ is an observation and $a_j$ is the corresponding expert action. The tokenizer $h: \mathcal{A} \to \{1, \dots, C\}^L$ maps an action chunk to a token sequence of length $L$. The VLM backbone $\pi_\theta$ defines an autoregressive distribution over token sequences conditioned on the observation:
\begin{equation}
    \pi_\theta(c \mid s) = \prod_{\ell=1}^{L} \pi_\theta(c_\ell \mid s, c_{<\ell}),
\end{equation}
where each factor $\pi_\theta(c_\ell \mid s, c_{<\ell})$ is a softmax over the vocabulary $\{1,\dots,C\}$.
The true conditional distribution is $P^*(c \mid s) = \Pr[h(a) = c \mid s]$, representing the probability that a demonstration action at observation $s$ produces token sequence $c$.

\noindent\textbf{Notation convention for the proofs below.} For notational compactness, we use $\pi_\theta(c \mid s)$ and $P^*(c\mid s)$ to denote the \emph{sequence-level} probabilities $\pi_\theta(h(a)\mid s)$ and $P^*(h(a)\mid s)$ throughout this section. All results hold under the autoregressive factorization above, since the NLL loss decomposes into per-token cross-entropies, each of which is a strictly proper scoring rule.

\subsection{Proper Scoring Rule Foundation}

\begin{definition}[Strictly proper scoring rule~\citep{gneiting2007strictly}]
\label{def:proper_scoring}
A scoring rule $S: \mathcal{P} \times \mathcal{Y} \to \mathbb{R}$ is \emph{strictly proper} if, for all distributions $P$ over $\mathcal{Y}$:
\begin{equation}
    \mathbb{E}_{y \sim P}[S(P, y)] > \mathbb{E}_{y \sim P}[S(q, y)] \quad \forall\, q \neq P.
\end{equation}
\end{definition}

\begin{proposition}[Cross-entropy as strictly proper scoring rule]
\label{prop:ce_proper}
The negative log-likelihood scoring rule $S(q, y) = \log q(y)$ is strictly proper. Consequently, the cross-entropy loss:
\begin{equation}
    \mathcal{L}_{\text{CE}}(\theta) = -\mathbb{E}_{(s,c) \sim \mathcal{D}_{\text{bucket}}}[\log \pi_\theta(c \mid s)]
\end{equation}
has a unique global minimizer (over all measurable functions) at $\pi_\theta(\cdot \mid s) = P^*(\cdot \mid s)$ for each $s$.
\end{proposition}

\begin{proof}
This is a classical result. By the information inequality (Gibbs' inequality), for any distributions $p$ and $q$ over $\{1, \dots, C\}$:
\begin{equation}
    -\sum_{c=1}^C p(c) \log p(c) \leq -\sum_{c=1}^C p(c) \log q(c),
\end{equation}
with equality if and only if $p = q$. Taking $p = P^*(\cdot \mid s)$, the expected cross-entropy $\mathbb{E}_s[-\sum_c P^*(c \mid s) \log \pi_\theta(c \mid s)]$ is uniquely minimized when $\pi_\theta(\cdot \mid s) = P^*(\cdot \mid s)$ for all $s$.
\end{proof}

\subsection{Verifier Soundness Theorem}

The proper scoring rule property has a direct consequence for action verification: as training converges, the bucket head's scoring function becomes consistent with the true quality ordering.

\begin{definition}[Calibration error]
\label{def:calibration}
The log-space calibration error of $\pi_\theta$ over a candidate set $\mathcal{A}=\{a_1,\ldots,a_M\}$ is:
\begin{equation}
    \varepsilon_{\mathrm{cal}}(\theta) \coloneqq \mathbb{E}_s\!\left[\max_{m\in\{1,\ldots,M\}} \left| \log\pi_\theta\bigl(h(a_m) \mid s\bigr) - \log P^*\bigl(h(a_m) \mid s\bigr) \right|\right],
\end{equation}
where $\pi_\theta(h(a_m)\mid s)$ and $P^*(h(a_m)\mid s)$ denote the sequence-level probabilities under the autoregressive factorization (Section~\ref{app:calibration_proof}, Setup and Notation).
\end{definition}

\begin{definition}[Quality gap]
\label{def:quality_gap}
For two candidates $a_m, a_{m'}$ at observation $s$, the \emph{log-probability quality gap} is:
\begin{equation}
    \gamma(a_m, a_{m'}, s) \coloneqq \log P^*\bigl(h(a_m) \mid s\bigr) - \log P^*\bigl(h(a_{m'}) \mid s\bigr).
\end{equation}
When $\gamma > 0$, candidate $a_m$ is ``better'' in the sense that its token-sequence probability under the expert distribution is higher.
\end{definition}

\begin{theorem}[Verifier soundness]
\label{thm:verifier_soundness}
Suppose the log-space calibration error satisfies $\varepsilon_{\mathrm{cal}}(\theta) < \gamma_{\min} / 2$, where $\gamma_{\min} = \min_{s, m \neq m': \gamma(a_m, a_{m'}, s) > 0} \gamma(a_m, a_{m'}, s)$ is the minimum positive log-probability quality gap. Then for all observations $s$ and all candidate pairs with $P^*(h(a_m) \mid s) > P^*(h(a_{m'}) \mid s)$:
\begin{equation}
    \log\pi_\theta\bigl(h(a_m) \mid s\bigr) > \log\pi_\theta\bigl(h(a_{m'}) \mid s\bigr),
\end{equation}
i.e., the verifier's log-probability ranking is consistent with the true quality ordering. Equivalently, $\text{score}(a_m) > \text{score}(a_{m'})$ under the scoring function defined in Section~\ref{sec:cotrain}.
\end{theorem}

\begin{proof}
By the triangle inequality and the definition of log-space calibration error:
\begin{align}
    &\log\pi_\theta(h(a_m) \mid s) - \log\pi_\theta(h(a_{m'}) \mid s) \notag\\
    &\quad\geq \bigl[\log P^*(h(a_m) \mid s) - \varepsilon_{\mathrm{cal}}\bigr]- \bigl[\log P^*(h(a_{m'}) \mid s) + \varepsilon_{\mathrm{cal}}\bigr] \\
    &\quad= \gamma(a_m, a_{m'}, s) - 2\varepsilon_{\mathrm{cal}} \\
    &\quad\geq \gamma_{\min} - 2\varepsilon_{\mathrm{cal}} > 0.
\end{align}
\end{proof}

\begin{remark}[Practical implications]
Theorem~\ref{thm:verifier_soundness} characterizes when estimation error is small enough to preserve the demonstration-likelihood ordering. Cross-entropy provides a population objective for learning this distribution, but finite-data optimization and distribution shift can prevent the required error bound from holding. The result is therefore a sufficient condition for ranking consistency; identifying likelihood with task utility requires the additional alignment assumption discussed in Section~\ref{sec:theory}.
\end{remark}

\section{Co-Training Synergy: Theoretical Perspective}
\label{app:synergy}

The empirical observation that co-training with the bucket classification loss accelerates flow-matching convergence (Figure~\ref{fig:co_training}) admits a natural explanation within the calibration framework.

\begin{proposition}[Multi-task regularization]
\label{prop:multi_task}
Let $f_\theta = (f_\theta^{\text{shared}}, f_\theta^{\text{fm}}, f_\theta^{\text{bucket}})$ denote the decomposition into shared backbone, flow-matching head, and bucket head. The co-training objective $\mathcal{L}_{\text{total}} = \mathcal{L}_{\text{fm}} + \beta \cdot \mathcal{L}_{\text{bucket}}$ implicitly regularizes the shared representation $f_\theta^{\text{shared}}$ by encouraging it to encode information sufficient for both action generation and action classification. Specifically, the bucket objective enforces that $f_\theta^{\text{shared}}(s)$ must distinguish observations where different action buckets are optimal---precisely the same discriminative structure needed for the flow-matching head to denoise toward the correct action mode.
\end{proposition}

This explains why the gradient cosine similarity between the two objectives is positive ($+0.28$ at the default $\beta{=}3$, Table~\ref{tab:multi_obj_kernel}): both objectives push the backbone toward representations that are informative about which action is appropriate for a given observation. The bucket objective, by requiring explicit mode discrimination through its classification structure, provides a stronger gradient signal early in training when the flow-matching loss landscape is relatively flat.

\subsection{Empirical Validation of Co-Training Dynamics}
\label{app:cotrain_dynamics}

We measure the gradient alignment and representation quality throughout co-training to validate the multi-task regularization hypothesis.

\begin{table}[h]
\centering
\caption{\textbf{Multi-objective training analysis.} All configurations use Qwen2.5-VL-3B trained for 64 epochs on LIBERO. $\|\Delta\theta\|/\|\theta_0\|$: relative parameter displacement. Cosine($\nabla_{\text{fm}}$, $\nabla_{\text{bucket}}$): average cosine similarity between the gradients of the two objectives during training.}
\label{tab:multi_obj_kernel}
\resizebox{0.85\linewidth}{!}{
\begin{tabular}{l c c c}
\toprule
\textbf{Training Configuration} & $\|\Delta\theta\| / \|\theta_0\|$ & \textbf{Cos}($\nabla_{\text{fm}}$, $\nabla_{\text{bucket}}$) & \textbf{LIBERO SR (\%)} \\
\midrule
Flow-matching Loss only & 0.0081 & -- & 91.9 \\
Bucket Loss only & 0.0063 & -- & -- \\
\midrule
Co-training ($\beta{=}1$) & 0.0092 & +0.31 & 95.4 \\
Co-training ($\beta{=}3$, default) & 0.0107 & +0.28 & 96.8 \\
Co-training ($\beta{=}5$) & 0.0126 & +0.24 & 96.2 \\
Co-training ($\beta{=}20$) & 0.0168 & +0.19 & 94.7 \\
\bottomrule
\end{tabular}
}
\end{table}

\vpara{Analysis.} Table~\ref{tab:multi_obj_kernel} reveals that the positive gradient cosine similarity between the two objectives ($+0.28$ at default $\beta{=}3$) confirms they push the backbone in partially aligned directions. This gradient coherence is the structural reason why co-training improves over single-objective training: the bucket objective provides an additional discriminative signal that accelerates the formation of action-relevant representations in the shared backbone. Excessive $\beta$ ($\geq 20$) degrades performance because the classification gradient dominates and the backbone overspecializes for bucket discrimination at the expense of smooth action generation.

\subsection{Expected Selection Error Bound}
\label{app:selection_error}

We derive an upper bound on the expected sub-optimality incurred by selecting a candidate action via the bucket probability verifier rather than the true Q-function.

\subsubsection{Setup}

Consider $M$ candidate actions $\mathcal{A} = \{a_1, \dots, a_M\}$ sampled from the diffusion expert at observation $s$. Let
\begin{equation}
    a^* = \arg\max_{a_m \in \mathcal{A}} Q(a_m \mid s)
\end{equation}
denote the oracle-best candidate (highest true Q-value), and
\begin{equation}
    \hat{a} = \arg\max_{a_m \in \mathcal{A}} \pi_\theta\bigl(h(a_m) \mid s\bigr)
\end{equation}
the verifier's selection (highest bucket probability). Define the per-candidate sub-optimality gap $\Delta_m = Q(a^* \mid s) - Q(a_m \mid s) \geq 0$.

\noindent\textbf{Note on $Q$.} The action-value function $Q(a \mid s)$ here denotes the oracle expected return (e.g., task success probability) and serves purely as an analytical reference for measuring sub-optimality. It is \emph{not} a learned component of our system---TokenPB does not train or deploy any Q-network. The bound quantifies how much regret is incurred by using the bucket probability $\pi_\theta(h(a) \mid s)$ as a proxy for $Q$.

\subsubsection{Crude Bound}

Let $q_m = \Pr[\hat{a} = a_m]$ denote the probability that the verifier selects candidate $m$ (accounting for stochasticity in the sampling process and possible ties). The expected selection error is:
\begin{equation}\label{eq:selection_error}
    \mathbb{E}\bigl[Q(a^* \mid s) - Q(\hat{a} \mid s)\bigr] = \sum_{m=1}^M q_m\, \Delta_m \;\leq\; \delta \cdot \Delta_{\max},
\end{equation}
where $\delta = \Pr[\hat{a} \neq a^*] = 1 - q_{m^*}$ is the misselection probability and $\Delta_{\max} = \max_m \Delta_m$ is the worst-case gap. This bound is loose because it ignores the structure of $q_m$---specifically, that the scoring monotonicity (Theorem~\ref{thm:scoring_monotonicity}) concentrates probability mass on high-Q candidates.

\subsubsection{Tight Bound via Gibbs Modeling}

To obtain a tighter characterization, we model the verifier's selection mechanism more precisely. The bucket log-probability $p_m = \log \pi_\theta(h(a_m) \mid s)$ serves as the scoring function. Since the $\arg\max$ selector is deterministic given $\{p_m\}$, we account for uncertainty by modeling the effective selection as a Gibbs distribution with inverse temperature $\lambda > 0$:
\begin{equation}
    q_m = \frac{\exp(\lambda \cdot p_m)}{\sum_{m'=1}^M \exp(\lambda \cdot p_{m'})}.
\end{equation}
This models the regime where the $\arg\max$ is ``soft''---appropriate when score differences are moderate relative to noise. The $\arg\max$ selector is recovered as $\lambda \to \infty$.

\begin{proposition}[Selection error upper bound]\label{prop:selection_error}
Under the Gibbs selection model with inverse temperature $\lambda$, and assuming the scoring monotonicity (Theorem~\ref{thm:scoring_monotonicity}) holds with calibration error bounded by $\varepsilon_{\mathrm{cal}}$, the expected selection error satisfies:
\begin{equation}\label{eq:selection_error_tight}
    \mathbb{E}\bigl[Q(a^* \mid s) - Q(\hat{a} \mid s)\bigr] \;\leq\; \frac{\log M}{\lambda} + L_Q \cdot \varepsilon_{\mathrm{cal}},
\end{equation}
where $L_Q$ is the Lipschitz constant of $Q$ with respect to the log-probability metric (i.e., the maximum Q-variation between candidates whose log-probabilities $p_m = \log\pi_\theta(h(a_m)\mid s)$ differ by $\varepsilon_{\mathrm{cal}}$).
\end{proposition}

\begin{proof}
The first term follows from the standard regret bound for Gibbs sampling. For any distribution $q$ over $M$ items with rewards $r_m$ and $q_m \propto \exp(\lambda \cdot r_m)$:
\begin{equation}
    \max_m r_m - \sum_m q_m r_m \leq \frac{\log M}{\lambda}.
\end{equation}
This is a direct consequence of the log-sum-exp inequality: $\max_m r_m \leq \frac{1}{\lambda}\log\sum_m \exp(\lambda r_m) \leq \max_m r_m + \frac{\log M}{\lambda}$.

In our setting, if $p_m$ were a perfect proxy for $Q(a_m \mid s)$ (up to affine transformation), this bound would apply directly with $r_m = Q(a_m \mid s)$. The scoring monotonicity guarantees that $p_m$ is monotonically related to the true quality $P^*(h(a_m) \mid s)$, which in turn correlates with $Q(a_m \mid s)$ through the expert demonstration distribution. The deviation from a perfect proxy arises from calibration error: when $\pi_\theta \neq P^*$, the bucket probability may mis-rank candidates by at most
\begin{equation}
    |p_m - p_m^{\text{ideal}}| \leq \varepsilon_{\mathrm{cal}}.
\end{equation}
The resulting additional regret from this perturbation is bounded by $L_Q \cdot \varepsilon_{\mathrm{cal}}$ (since the Q-function varies by at most $L_Q$ per unit of log-probability perturbation), yielding the second term.
\end{proof}

\subsubsection{Interpretation}

The selection bound separates the effect of a finite-temperature selector from estimation error. The $\log M/\lambda$ term decreases as the selector becomes sharper for a fixed candidate set, while the calibration term captures the effect of imperfect likelihood estimates. This interpretation depends on the assumed relationship between demonstration likelihood and task utility; the likelihood objective alone does not establish that relationship.

\section{Experimental Settings}
\label{sec:exp}
Experimental settings mainly include the key parameters of the action expert model, training configurations, and the evaluation criteria for real-world experiments.
\subsection{Action Expert Model}\label{action}
{\renewcommand{\arraystretch}{1.3} 
\begin{table}[t]
\centering
\caption{\textbf{Action Expert Model Parameters.}}
\label{tb:parameter}
\begin{tabular}{l|lc}
\toprule
\textbf{Category} & \textbf{Parameter} & \textbf{Value} \\
\midrule
\multirow{4}{*}{\textbf{Model Size}} & Number of layers & 16 \\
& Hidden size & 1536 \\
& Number of attention heads & 32 \\
& Parameter size & $\sim$500M\\
\midrule
\multirow{2}{*}{\textbf{Action Config}}& De-noising step & 4\\
& Number of candidate action chunks $M$  & 16\\
\midrule
\multirow{3}{*}{\textbf{Layer Structure}} & Self-attention layer & All \textbf{even} layers \\
& Cross-attention layer & All \textbf{odd} layers \\
& Dropout & 0.2 \\
\midrule
\textbf{Activation} & Activation function & GELU \\
\bottomrule
\end{tabular}
\label{9}
\end{table}
} 
Inspired by GR00T-N1, our action expert utilizes an alternating sequence of DiT and self-attention during inference, as detailed in Table~\ref{9}. To balance training efficiency with strong generalization, we implemented a 16-layer Transformer framework where self-attention and cross-attention layers are interleaved. With a total parameter count of roughly 500M, the model is trained entirely from scratch using random weights for each experiment. 
\subsection{Training Settings}\label{training}
\begin{table}[t]
\centering
\caption{Training Configuration on LIBERO, CALVIN, RoboTwin2.0 and Real-world benchmark.}
\resizebox{\linewidth}{!}{
\renewcommand{\arraystretch}{1.3}
\begin{tabular}{cl|cccc}
\toprule
\textbf{Category} & \textbf{Parameter}  & \textbf{Real-World} & \textbf{LIBERO} & \textbf{CALVIN} & \textbf{RoboTwin2.0\textsuperscript{*}} \\
\addlinespace[0.5em]
\hline
\multirow{5}{*}{\textbf{Dataset Statistics}} & Total episodes & 2005 & 1693 & 17870 & 2500 \\
& Total frames & 366152 & 273465 & 1071743 & 549787 \\
& Task & 4 & 40 & 389 & 50\\
& Instruction type  & 741 & 40 & 389 & 2420\\
& FPS & 5 & 10 & 10 & 50\\
\hline
\multirow{4}{*}{\textbf{Training Parameters}} & Batch size & 144 & 144 & 144 & 120\\
& Epoch & 32 & 64 & 16 & 96\\
& Learning rate & 5$\times10^{-5}$ & 5$\times10^{-5}$ & 5$\times10^{-5}$ & 2$\times10^{-5}$\\
& Weight decay & 1$\times10^{-5}$ & 1$\times10^{-5}$ & 1$\times10^{-5}$ & 1$\times10^{-5}$\\
\hline
\multirow{4}{*}{\textbf{Other Settings}} & Seed & 42 & 42 & 42 & 42\\
& Discrete tokenizer & Textual FAST & Textual FAST & Textual FAST & Textual FAST\\
& Action chunk length $H$ & 5 & 16 & 16 & 32\\
& Co-training loss coefficient $\beta$ & 3 & 3 & 3 & 1\\
\bottomrule
\end{tabular}
}
\par\smallskip
{\footnotesize $^*$ In actual training, fine-tuning is performed on every single task, consistent with the official settings.\par}
\label{12}
\end{table}
Our experiments are primarily conducted on LIBERO, RoboTwin2.0, and in real-world environments. For each benchmark, we use its corresponding dataset to co-train the action expert for continuous actions and the VLM for discrete action tokens. For the evaluation stage, we test the performance of the three inference schemes---auto-regression, diffusion, and the TokenPB framework---using the same checkpoint. Refer to Table \ref{12} for specific training settings.

\subsection{Complete TokenPB Workflow}\label{app:algorithm}

We present the full TokenPB pipeline in Algorithm~\ref{alg:tokenpb_train} (co-training) and Algorithm~\ref{alg:tokenpb_infer} (inference with verification).

\begin{algorithm}[h]
\caption{TokenPB Co-Training}
\label{alg:tokenpb_train}
\begin{algorithmic}[1]
\REQUIRE Training set $\mathcal{D} = \{(s_j, a_j)\}_{j=1}^n$; VLM backbone $f_\theta$; diffusion action expert $g_\phi$; tokenizer $h$; token vocabulary size $C$; co-training loss coefficient $\beta$
\ENSURE Co-trained parameters $(\theta^*, \phi^*)$
\STATE Initialize VLM from pretrained weights $\theta_0$; initialize action expert randomly $\phi_0$
\FOR{each mini-batch $\{(s, a^{\text{demo}})\} \sim \mathcal{D}$}
    \STATE $f_c \leftarrow f_\theta(s)$ \hfill $\triangleright$ observation-conditioned VLM feature
    \STATE $c^{\text{demo}} = (c_1,\ldots,c_L) \leftarrow h(a^{\text{demo}})$ \hfill $\triangleright$ tokenize demo action into sequence of length $L$
    \STATE $\mathcal{L}_{\text{bucket}} \leftarrow -\frac{1}{L}\sum_{\ell=1}^{L} \log \pi_\theta(c_\ell \mid s, c_{<\ell})$ \hfill $\triangleright$ autoregressive NLL over token sequence
    \STATE Sample noise $\epsilon \sim \mathcal{N}(0, I)$, timestamp $\tau \sim \text{Uniform}(0, 1)$
    \STATE $a^\tau \leftarrow \tau \cdot a^{\text{demo}} + (1-\tau) \cdot \epsilon$ \hfill $\triangleright$ noisy action at time $\tau$
    \STATE $\mathcal{L}_{\text{fm}} \leftarrow \| g_\phi(a^\tau, \tau, f_c) - (a^{\text{demo}} - \epsilon) \|^2$ \hfill $\triangleright$ flow-matching objective
    \STATE $\mathcal{L}_{\text{total}} \leftarrow \mathcal{L}_{\text{fm}} + \beta \cdot \mathcal{L}_{\text{bucket}}$
    \STATE Update $(\theta, \phi) \leftarrow (\theta, \phi) - \eta \nabla_{(\theta,\phi)} \mathcal{L}_{\text{total}}$
\ENDFOR
\RETURN $(\theta^*, \phi^*)$
\end{algorithmic}
\end{algorithm}

\begin{algorithm}[h]
\caption{TokenPB Inference: Multi-Candidate Verification}
\label{alg:tokenpb_infer}
\begin{algorithmic}[1]
\REQUIRE Observation $s_t$; co-trained VLM $f_{\theta^*}$; diffusion action expert $g_{\phi^*}$; number of candidates $M$
\ENSURE Selected action chunk $\hat{a}^*$
\STATE \textcolor{gray}{\textit{// Stage 1: Draft --- generate $M$ candidate action chunks}}
\STATE $f_c \leftarrow f_{\theta^*}(s_t)$ \hfill $\triangleright$ single VLM forward pass for observation feature
\FOR{$m = 1, \dots, M$}
    \STATE Sample $\epsilon_m \sim \mathcal{N}(0, I)$ \hfill $\triangleright$ independent noise seeds
    \STATE $a_m \leftarrow g_{\phi^*}(\epsilon_m, f_c)$ \hfill $\triangleright$ 4-step Euler ODE integration
\ENDFOR
\STATE $\mathcal{A}_{s_t} \leftarrow \{a_1, \dots, a_M\}$
\STATE
\STATE \textcolor{gray}{\textit{// Stage 2: Verify --- score candidates via token-sequence probability}}
\FOR{$m = 1, \dots, M$}
    \STATE $c^{(m)} = (c^{(m)}_1,\ldots,c^{(m)}_L) \leftarrow h(a_m)$ \hfill $\triangleright$ tokenize candidate
    \STATE $\text{score}(a_m) \leftarrow \frac{1}{L}\sum_{\ell=1}^{L} \log \pi_{\theta^*}(c^{(m)}_\ell \mid s_t, c^{(m)}_{<\ell})$ \hfill $\triangleright$ teacher-forcing scoring
\ENDFOR
\STATE $\hat{a}^* \leftarrow \arg\max_{a_m \in \mathcal{A}_{s_t}} \text{score}(a_m)$
\RETURN $\hat{a}^*$
\end{algorithmic}
\end{algorithm}

\vpara{Key design notes.} (1)~The co-training loss $\mathcal{L}_{\text{total}}$ jointly trains both the VLM backbone and the action expert via a shared hidden feature $f_c$, allowing the bucket classification objective to shape action-relevant representations that also benefit flow-matching generation. (2)~At inference time, all $M$ candidates share the same observation prefix, so their token sequences can be scored in a single batched forward pass via teacher-forcing; the verification overhead is therefore nearly constant with respect to $M$. (3)~The total inference cost is approximately two VLM forward passes (one for drafting, one for verification) plus $M$ lightweight flow-matching sampling steps (4-step Euler integration each), making TokenPB nearly constant-cost with respect to $M$.

\subsection{Real-World Experiment Scoring Details}\label{metric}
For each rollout, we record a normalized success rate \(S \in [0,1]\), representing the percentage of task completion.
This metric accounts for partial completion, which is common in real-world manipulation, and is more informative than binary success indicators.
To ensure clarity, we report the success rate as \(\text{Success Rate} = 100\% \times S\) in the main paper, while maintaining \(S\in[0,1]\) in the appendix tables.
\begin{itemize}[leftmargin=1.1em]
    \item \textbf{Push Blocks}: In the block-pushing task, the total score is comprised of a partial score and a completion score. For example, in the "push CAT forward" task, pushing the "C", "A", and "T" blocks accounts for 25\% of the partial score each. The remaining 25\% is awarded if all three blocks are successfully pushed and no other blocks are moved. \\
    For each block-pushing trial, we assign an individual score \(s \in [0,1]\)  based on the final block configuration, using the following rubric:
    \begin{itemize}
        \item \(s = 0.0\): The current block remains untouched throughout the experiment.
        \item \(s = 0.3\): The current block is touched but not successfully pushed.
        \item \(s = 0.5\): The current block is pushed but did not reach the designated finish line.
        \item \(s = 1.0\): The current block is successfully pushed past the finish line.
    \end{itemize}
    We define $s^r$ as the aggregate score for the $N$ target blocks in the instructions, and $s^w$ as the penalty score for the $M$ unmentioned blocks. Ideally, an optimal trajectory should successfully push all mentioned blocks while leaving the unmentioned blocks untouched, leading to the following scoring criteria:
    \[S_{\text{push}} = \frac{1}{N+1}(\sum_{i=1}^{N} s^r_i+1-\frac{1}{M}\sum_{i=1}^{M} s^w_i).\]
    \item \textbf{Clean Table}: The table-cleaning task is categorized into three distinct phases: opening the drawer, placing all items (typically three) from the table into the drawer sequentially, and closing the drawer.  
    
    We define the score for opening and closing a drawer as binary values, $s_o, s_c\in \{0,1\}$, where $1$ indicates the drawer was successfully manipulated, and $0$ indicates manipulation failure or no attempt. For each pick-up attempt, we assign an individual nesting score $s \in [0,1]$ using the following rubric based on the final block configuration:
    \begin{itemize}
        \item \(s = 0.0\): The current object remains untouched throughout the experiment.
        \item \(s = 0.8\): The current object is picked up but not successfully placed in the drawer.
        \item \(s = 1.0\): The current object is successfully picked up and placed in the drawer.
    \end{itemize}
    We denote $s^d$ as the score for the $N$ objects on the table. The final score for the table-cleaning task is calculated as follows:
    \[S_{\text{clean}} = \frac{1}{N+2}(s_o+s_c+\sum_{i=1}^{N} s^d_i)\]
    \item \textbf{Pick \& Place}: In the pick-and-place task, the total score is comprised of a 60\% completion component and a 40\% accuracy component. The completion score $s^d$ is awarded based on the successful grasping and placement of the $N$ target objects specified in the instruction, following the same protocol as the table-cleaning task. The accuracy score $s^r$ refers to the proportion of $N$ correct objects identified by the model among a total of $M$ objects. Notably, attempts are counted towards the accuracy score even if the grasp ultimately fails.
    we define score $s\in\{0,1\}$ based on whether the object is identified or not, For each object j, define $r_j=1$ if it is mentioned; define $a_j$=1 if the agent attempts to grasp it at least once.
    \[S_{\text{pick}} = 60\% \times \frac{1}{N}\sum_{i=1}^{N} s^d_i+40\% \times (1-\frac{1}{M}\sum_{j=1}^{M} [a_j\neq r_j])\]
    \item \textbf{Hang Cups}: The ``hang cups'' task consists of two primary processes: picking up the cup and hanging the cup. To better evaluate the model's sensitivity to instruction-specified orientations, we decompose the task into five sequential components: moving above the corresponding target $s_m$, successfully grasping the cup $s_g$, moving to the vicinity of the hook in the corresponding orientation $s_h$, aligning the cup handle with the hook $s_a$, and releasing the gripper to successfully hang the cup $s_r$:
    \[S_{\text{cup}} = \frac{1}{5} \times(s_m+s_g+s_h+s_a+s_r)\]
    
\end{itemize}

\subsection{Statistical Analysis of Real-World Results}
\label{sec:realworld_stats}

To quantify the reliability of our real-world evaluation, we report the number of evaluation rollouts per task, standard deviations, 95\% bootstrap confidence intervals, and paired $t$-test $p$-values comparing TokenPB against the single-sample baseline (from the same co-trained checkpoint). For the in-distribution setting, each instruction variant is evaluated with 3 independent rollouts under randomized initial object configurations. For the unseen setting, we increase the evaluation density to 18 rollouts per instruction to obtain more reliable estimates under distribution shift. The total rollout counts are shown in Tables~\ref{tab:realworld_stats} and~\ref{tab:realworld_unseen_stats}.

\begin{table}[h]
\centering
\caption{\textbf{Statistical analysis of Real-World (In-Distribution) results.} $n$: total evaluation rollouts per task. $\mu \pm \sigma$: mean success rate (\%) $\pm$ standard deviation across rollouts. CI: 95\% bootstrap confidence interval (1000 resamples). $p$: paired $t$-test $p$-value comparing TokenPB vs.\ single-sample baseline. All models use Qwen2.5-VL-3B.}
\label{tab:realworld_stats}
\resizebox{\linewidth}{!}{
\begin{tabular}{l c c c c c}
\toprule
\textbf{Task} & $n$ & \textbf{Baseline ($\mu \pm \sigma$)} & \textbf{TokenPB ($\mu \pm \sigma$)} & \textbf{TokenPB 95\% CI} & $p$\textbf{-value} \\
\midrule
Push Blocks & 33 & $56.2 \pm 18.4$ & $75.4 \pm 15.7$ & [70.1, 80.8] & $< 0.001$ \\
Clean Table & 24 & $52.5 \pm 21.6$ & $72.5 \pm 17.3$ & [65.4, 79.6] & $< 0.005$ \\
Pick \& Place & 30 & $45.5 \pm 22.1$ & $75.0 \pm 16.8$ & [68.8, 81.2] & $< 0.001$ \\
Hang Cups & 24 & $43.7 \pm 5.8$ & $62.5 \pm 5.2$ & [60.4, 64.6] & $< 0.001$ \\
\midrule
\textbf{Overall} & \textbf{111} & $49.5 \pm 18.7$ & $71.4 \pm 15.1$ & [68.6, 74.2] & $< 0.001$ \\
\bottomrule
\end{tabular}
}
\end{table}

\begin{table}[h]
\centering
\caption{\textbf{Statistical analysis of Real-World (Unseen) results.} Each of the 8 unseen instructions is evaluated with 18 independent rollouts ($n = 144$ total). All models use Qwen2.5-VL-3B.}
\label{tab:realworld_unseen_stats}
\resizebox{\linewidth}{!}{
\begin{tabular}{l c c c c c}
\toprule
\textbf{Task} & $n$ & \textbf{Baseline ($\mu \pm \sigma$)} & \textbf{TokenPB ($\mu \pm \sigma$)} & \textbf{TokenPB 95\% CI} & $p$\textbf{-value} \\
\midrule
Push Blocks & 54 & $5.0 \pm 8.7$ & $45.0 \pm 20.4$ & [39.5, 50.5] & $< 0.001$ \\
Clean Table & 18 & $8.0 \pm 13.8$ & $24.0 \pm 18.2$ & [15.5, 32.5] & $< 0.01$ \\
Pick \& Place & 54 & $28.0 \pm 19.5$ & $36.0 \pm 17.8$ & [31.3, 40.7] & $< 0.01$ \\
Hang Cups & 18 & $32.0 \pm 5.1$ & $40.0 \pm 4.5$ & [37.9, 42.1] & $< 0.001$ \\
\midrule
\textbf{Overall} & \textbf{144} & $18.3 \pm 16.8$ & $36.3 \pm 18.9$ & [33.2, 39.4] & $< 0.001$ \\
\bottomrule
\end{tabular}
}
\end{table}

As shown in Tables~\ref{tab:realworld_stats} and~\ref{tab:realworld_unseen_stats}, the in-distribution real-world improvements are statistically significant across all four tasks (all $p < 0.001$), with non-overlapping 95\% confidence intervals between the TokenPB and baseline conditions. For the unseen environment, the overall improvement (+18.0 points) achieves strong statistical significance ($p < 0.001$) with 144 total rollouts, and all four task categories reach significance individually (all $p < 0.01$). Across both settings, the Hang Cups task exhibits substantially lower variance ($\sigma \approx 5$) than the other tasks, reflecting its more constrained action space and deterministic success criteria---the cup either hooks onto the rack or does not, leaving little room for partial-credit variability. This low variability yields tight confidence intervals and strong significance even with fewer rollouts. We note that the relatively large standard deviations observed in Push Blocks, Clean Table, and Pick \& Place (9--22\%) are characteristic of real-world robotic manipulation evaluations where environmental stochasticity (object pose variation, gripper slip, lighting changes) contributes substantial trial-to-trial variability. Despite this inherent noise, the consistent improvement direction across all tasks and the aggregate statistical significance support the robustness of the reported gains.

\section{Task List of Real-World Tasks}\label{sec:realworldtasks}
\begin{figure*}[t]
    \centering
    \centerline{\includegraphics[width=\textwidth]{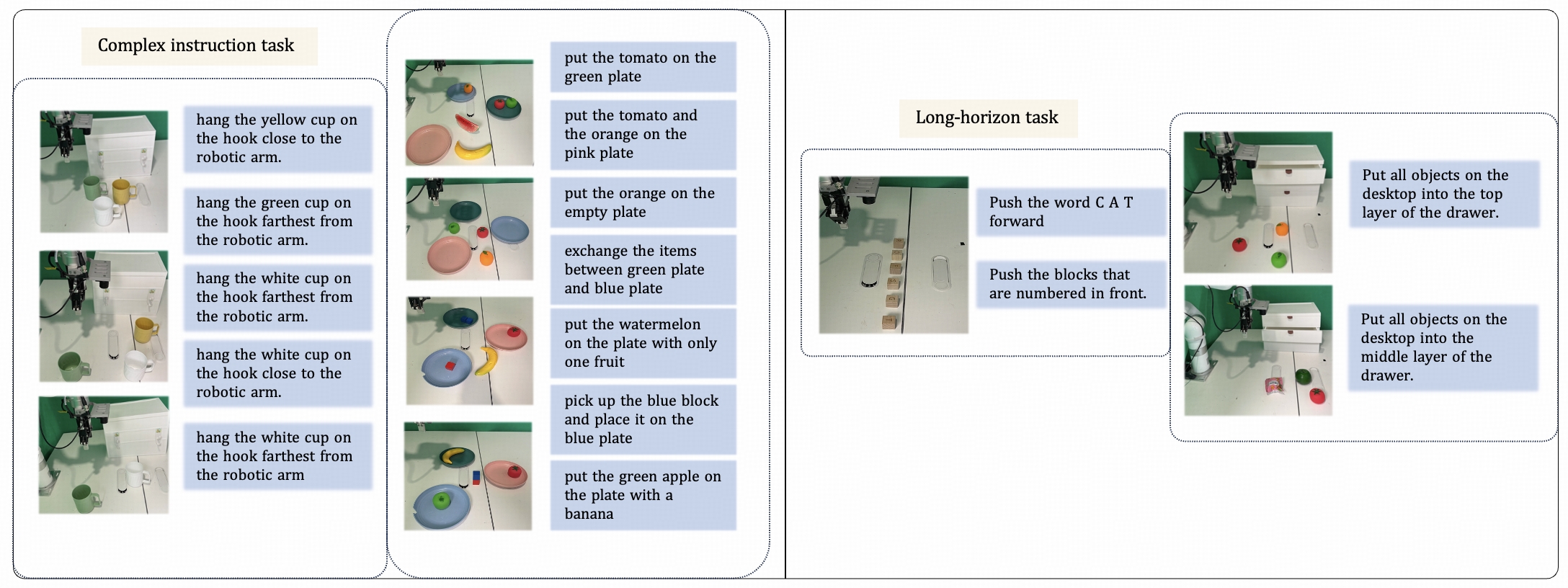}}
    \caption{\textbf{Partial real machine environment and instruction diagram.} Our testing is based on the xArm robotic arm and DaHuan gripper. In real-world we collect 2000 trajectories and select for testing in this work.}
    \label{fig7:real}
\end{figure*}
\label{real-world}
We collected over 2,000 real-world trajectories involving more than 50 distinct objects in a physical robot environment. Notably, certain task configurations were excluded to serve as held-out scenarios for evaluation. For the Unseen tasks, neither the instructions nor the environments were present in the training set.\\
Our primary test environment is illustrated in Figure~\ref{fig7:real}. The evaluation includes 34 distinct instruction types; to ensure representative results, some instructions were sampled multiple times in proportion to their natural frequency in the dataset.
\subsection{Push Blocks Task}
\begin{itemize}[leftmargin=1.1em]
\item Push the word C A T forward.
\item Push the word G L M forward.
\item Push the word D O G forward.
\item Push the word V L A forward.
\item Push the word R O S forward.
\item Push the word I C E forward.
\item Push the word Z H I P U forward.
\item Push the word F L Y forward.
\item Push the word L I K E forward.
\item Push the word B O X forward.
\item Push the blocks that are numbered in front.
\end{itemize}
\subsection{Clean Table Task}
\begin{itemize}[leftmargin=1.1em]
\item Put all objects on the desktop into the top layer of the drawer.
\item Put all objects on the desktop into the middle layer of the drawer. 
\end{itemize}
\subsection{Pick \& Place Task}
\begin{itemize}[leftmargin=1.1em]
\item Put the tomato on the green plate.
\item Put the tomato and the orange on the pink plate.
\item Put the orange on the empty plate.
\item exchange the items between the green plate and the blue plate.
\item Put the green apple on the plate with a banana.
\item Pick up the blue block and place it on the blue plate.
\item Put the banana on the plate with two blue blocks.
\item Put all items on the table into the pink plate.
\item Put the orange on the plate with a red block.
\item Put the watermelon on the plate with only one fruit.
\end{itemize}
\subsection{Hang Cups Task}
\begin{itemize}[leftmargin=1.1em]
\item Hang the white cup on the hook farthest from the robotic arm. 
\item Hang the white cup on the hook close to the robotic arm. 
\item Hang the yellow cup on the hook close to the robotic arm. 
\item Hang the green cup on the hook farthest from the robotic arm.
\end{itemize}
\subsection{Unseen Task}
\begin{itemize}[leftmargin=1.1em]
\item Exchange the position of the tomato and the pear.
\item Put the tomato on the plate that has only one item of fruit.
\item put as more as possible fruit to the green plate.
\item Push the blocks with 7 multiplied by 6 forward.
\item Push ahead the two blocks whose sum is over 14.
\item Push the blocks that have numbers on them forward.
\item Hang the green cup on the hook close to the robotic arm.
\item Put all the blue blocks into the top layer of the drawer.
\end{itemize}

\section{Four Action Tokenization Methods}
\label{appendix:token}
\begin{figure*}[t]
    \centering
    \centerline{\includegraphics[width=\textwidth]{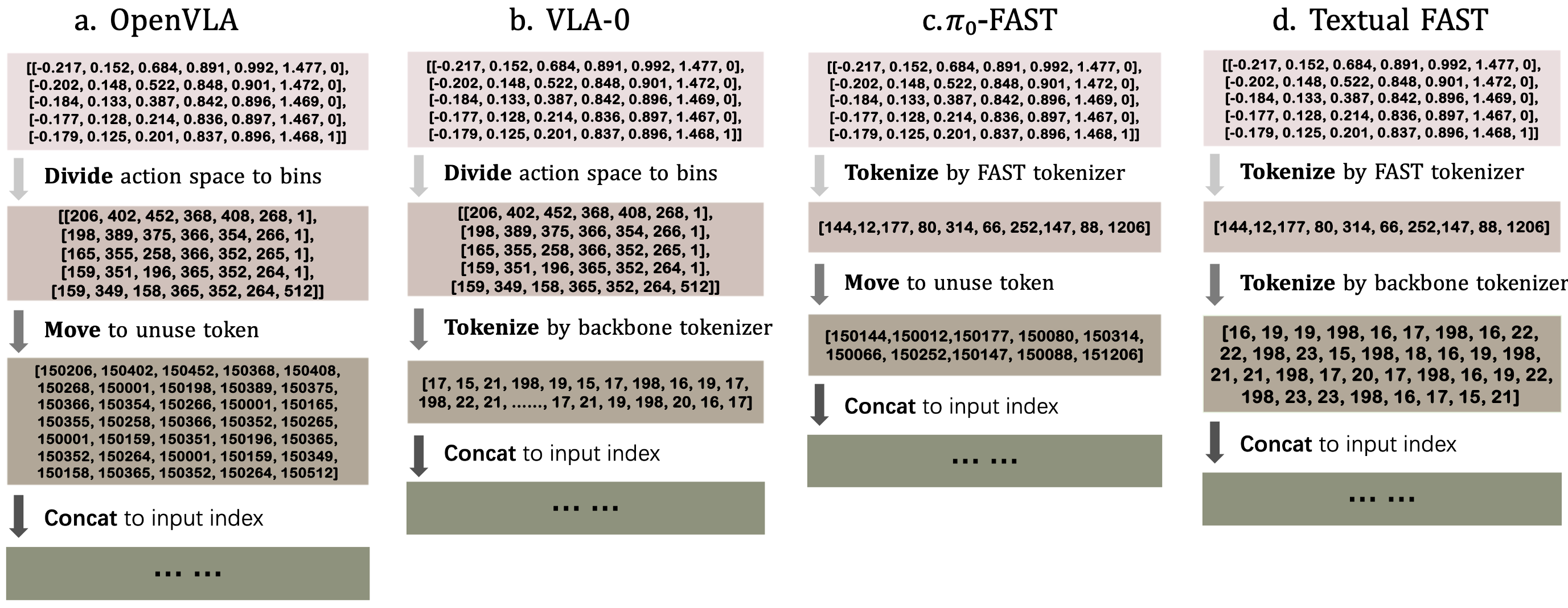}}
    \caption{\textbf{Examples of four discrete action token representation methods.} Assume VLM tokenizer encode `\textbackslash n' to 198, `0' to 14, `1' to 15, `2' to 16 and so on.}
    \label{fig8:token}
\end{figure*}
In this section, we illustrate the tokenization differences among four discrete encoding methods, using a concrete example where \textbf{action chunk}=5 and \textbf{action dim}=7 (see Figure \ref{fig8:token}). 
\subsection{Action Bins}
The Action Bins method, adopted by RT-2 and OpenVLA, determines range boundaries using the 1st and 99th percentiles of every dimension of action space and divides each action dimension into 512 bins. Although simple and intuitive, such methods perform poorly when learning dexterous skills with high-frequency control. Highly correlated action tokens diminish the effectiveness of the next token prediction objective used in auto-regressive VLAs. Intuitively, in such cases low token prediction loss can often be achieved with mappings as trivial as simply copying the most recent action token, leaving models in poor local optima~\cite{pertsch2025fast}.
\subsection{FAST}
The FAST tokenizer transforms continuous robot actions into tokens through the following steps:\\
\textbf{Quantile Normalization}: Raw action chunks are normalized such that the 1st and 99th quantiles of the training data map to $[-1, 1]$. This ensures robustness to outliers and facilitates cross-embodiment training.\\
\textbf{Discrete Cosine Transform (DCT)}: Each action dimension is converted into the frequency domain via DCT, transforming time-series signals into frequency coefficients.\\
\textbf{Quantization}: DCT coefficients are scaled by a hyperparameter $\gamma$ and rounded to integers. This creates a sparse matrix where high-frequency components often reduce to zero, effectively compressing the signal.\\
\textbf{Flattening}: The sparse coefficient matrix is flattened into a one-dimensional vector of integers using a "column-first" order. By prioritizing low-frequency components across dimensions, this order improves the stability of auto-regressive prediction.\\
\textbf{Byte Pair Encoding (BPE)}: The flattened sequence is further compressed into a shorter sequence of dense tokens using a trained BPE tokenizer. This step losslessly merges frequent coefficient combinations and eliminates redundant zero-valued components.\\
In the introduction to Textual FAST, we refer to this entire process simply as \textbf{FAST tokenization}.
\subsection{VLA-0}
Common VLM tokens and VLA action tokens often follow distinct distributions, which can lead to interference during the training process. VLA-0 proposes a simple method to circumvent this issue. First, VLA-0 divides each action dimension into several bins. Then, VLA-0 treats the sequence of actions as strings and encodes the action strings using VLM's tokenizer. In this way, the action sequence is processed similarly to numerical tasks in natural language processing, thereby reducing the modality gap. This optimization effectively accelerates convergence speed for VLAs that lack prior pre-training on discrete action tokens.
\subsection{Textual FAST}
\label{app:textualfast}
Building upon VLA-0, Textual FAST replaces dividing each action dimension into several bins with processing using FAST tokenization. Specifically, Textual FAST first processes actions into action tokens via FAST tokenization, and subsequently treats the sequence of actions as a string and encodes this action string using the VLM's tokenizer.

\vpara{Prediction architecture.} The compressed FAST representation is serialized as numerical text and encoded with the VLM's existing tokenizer. Its tokens are entries in the pretrained vocabulary, so the original language-modeling head predicts the bucket-token sequence directly; no additional prediction head or vocabulary expansion is required. Here, ``bucket head'' refers to the existing language-modeling head trained with the auxiliary bucket-token objective. This design reuses the pretrained token representations and can be integrated into the evaluated VLM--action-expert architectures without changing their prediction heads. The co-training convergence results in Figure~\ref{fig:co_training} and the tokenization comparison below support this choice.

\vpara{Score normalization.} We use the mean token log-likelihood in Section~\ref{sec:cotrain}, dividing each candidate's sequence log-likelihood by its own token count. This reduces the direct dependence of the summed log-likelihood on sequence length. It is the implemented ranking rule, rather than an additional normalization of the scores across candidates.

\vpara{Empirical comparison.} To quantify the impact of the tokenization scheme on downstream verification performance, we train Qwen2.5-VL with each of the four methods under identical settings and evaluate on LIBERO with TokenPB ($M{=}16$). Table~\ref{table:token_set} summarizes the results. Among the four schemes, Textual FAST achieves the highest average success rate (96.8\%), outperforming Action Bins (93.0\%), FAST (94.4\%), and VLA-0 (94.9\%) by clear margins.

Two complementary factors account for the advantage of Textual FAST. First, it inherits the strengths of both FAST and VLA-0: the DCT-based compression from FAST produces compact, low-redundancy frequency-domain representations, while the textual encoding from VLA-0 bridges the modality gap between action tokens and the VLM's pretrained vocabulary, facilitating more efficient co-training. Second, compared with Action Bins, Textual FAST distributes actions across a substantially larger number of hash buckets with lower per-bucket collision rates, yielding higher inter-bucket independence. This finer-grained and more orthogonal bucket structure enables the VLM to learn sharper probability distributions over action tokens, which in turn provides a more discriminative scoring signal for test-time verification.

Based on these results, we adopt Textual FAST as the default tokenization method in all experiments reported throughout this paper.

\begin{table}[t]
\centering
\setlength{\tabcolsep}{6pt}
\caption{\textbf{Impact of tokenization methods on LIBERO (\%).} All variants use Qwen2.5-VL-3B with TokenPB ($M{=}16$). The first row (Qwen2.5-VL-Diffusion) reports the single-sample baseline without verification. $\uparrow$/$\downarrow$ indicate improvement/degradation relative to this baseline.}
\resizebox{0.7\linewidth}{!}{
\begin{tabular}{l c c c c | c}
\toprule
\textbf{LIBERO}  & \thead{Spatial} & \thead{Object} & \thead{Goal}      & \thead{Long}      & \thead{Avg.} \\
\midrule
Qwen2.5-VL-Diffusion & 94.6       & 97.2      & 93.8         & 90.2              & 93.9\\
\midrule
Action Bins         & 92.8            & 95.6           & 94.4               & 89.0              & 93.0 $\downarrow$ \\
FAST         & 95.0            & 97.2           & 93.6              & 91.8              & 94.4 $\uparrow$ \\
VLA-0         & \textbf{96.8}            & 95.4           & 96.0              & 91.4              & 94.9 $\uparrow$ \\
\textbf{Textual FAST}        & 96.6            & \textbf{98.8}           & \textbf{97.0}              & \textbf{94.6}              & \textbf{96.8}  $\uparrow$\\
\bottomrule
\end{tabular}
}
\vspace{1ex}
\label{table:token_set}
\end{table}

\section{Full Test Results on LIBERO and RoboTwin2.0}\label{simulation}
\begin{table*}[t]
\setlength{\abovetopsep}{5pt}
\centering
\caption{\textbf{Test results on the LIBERO dataset.} ``Model-FAST'': FAST tokenizer-assisted auto-regression model. ``Model-GR00T'' and ``Model-GR00T+TokenPB'' share the same co-trained checkpoint (with both bucket classification and diffusion objectives) and differ only at inference: GR00T uses a single diffusion sample ($M{=}1$), while GR00T+TokenPB applies multi-candidate verification ($M{=}16$). Action chunk size per inference of our model is 16.}
\begin{tabular}{l c c c c c}
\toprule
Models   & \thead{Spatial} & \thead{Object} & \thead{Goal}      & \thead{Long}      & \thead{Avg.} \\
\midrule
Diffusion Policy~\cite{chi2025diffusion} & 78.3            & 92.5           & 68.3              & 50.5              & 72.4  \\
HybridVLA~\cite{liu2025hybridvla} & 93.6            & 90.6           & 90.2              & 84.4              & 90.2  \\
FlowVLA~\cite{zhong2025flowvla}  & 93.2            & 95.0           & 91.6              & 72.6              & 88.1  \\
SmolVLA~\cite{shukor2025smolvla}            & 93.0            & 94.0           & 91.0              & 77.0              & 88.8  \\
OpenVLA-OFT~\cite{kim2025fine}              & \ul{97.6}            & 98.4           & \tb{97.9}              & 94.5         & \ul{97.1}   \\
UniVLA~\cite{bu2025univla}    & 95.4      & \ul{98.8}      & 93.6         & 94.0         & 95.5  \\
Octo~\cite{team2024octo}             & 78.9            & 85.7           & 84.6              & 51.1              & 75.1 \\
OpenVLA~\cite{kim2025openvla}            & 84.7            & 88.4           & 79.2              & 53.7              & 76.5 \\
$\pi_{0}$-FAST~\cite{pertsch2025fast, goyal2025vla}              & 90.0            & 86.0           & 95.0              & 73.0              & 86.0\\
VLA-0~\cite{lee2025molmoact}         & 94.4            & 97.6           & 93.0              & 90.6              & 93.9  \\
GR00T-N1~\cite{bjorck2025gr00t}           & 94.4            & 97.6           & 93.0              & 90.6         & 93.9 \\
$\pi_{0}$~\cite{black2410pi0}             & 96.8            & \ul{98.8}      & 95.8              & 85.2              & 94.2 \\
$\pi_{0.5}$-KI~\cite{driess2025knowledge,lee2025molmoact}             & \tb{98.0}       & 97.8           & 95.6              & 85.8              & 94.3 \\
Discrete Diffusion VLA~\cite{kim2025fine}              & 97.2       & 98.6      & \ul{97.4}         & 92.0         & 96.3  \\
\midrule
Qwen2.5-VL-FAST          & 90.2       & 88.6      & 96.2         & 76.4              & 87.9  \\
Qwen2.5-VL-GR00T          & 94.6       & 97.2      & 93.8         & 90.2              & 93.9  \\
\tb{Qwen2.5-VL-GR00T+TokenPB}          & 96.6       & \ul{98.8}      & 97.0         & \ul{94.6}    & 96.8  \\
InternVL3.5-FAST          & 89.6       & 92.8      & 93.6         & 82.8              & 89.6  \\
InternVL3.5-GR00T        & 95.8       & 96.0      & 94.2         & 92.8              & 94.7  \\
\tb{InternVL3.5-GR00T+TokenPB}          & \ul{97.6}       & \tb{99.4}      & 97.2         & \tb{96.8}     & \tb{97.8}  \\
\bottomrule
\end{tabular}
\label{table:LIBERO}
\end{table*}

\begin{table}[t]
  \centering
  \footnotesize
  \setlength{\tabcolsep}{8pt}
  \caption{\textbf{RoboTwin 2.0 Simulation Benchmark.}}
  \resizebox{\linewidth}{!}{%
  \begin{tabular}{l*{12}{c}}
    \toprule
    \textbf{\makecell[c]{Simulation Task}} 
      & \multicolumn{2}{c}{\textbf{RDT}} 
      & \multicolumn{2}{c}{\textbf{Pi0}} 
      & \multicolumn{2}{c}{\textbf{ACT}} 
      & \multicolumn{2}{c}{\textbf{DP}} 
      & \multicolumn{2}{c}{\textbf{TokenPB}}
      & \multicolumn{2}{c}{\textbf{Qwen-Dif}} \\
    & \textbf{Easy} & \textbf{Hard} 
    & \textbf{Easy} & \textbf{Hard} 
    & \textbf{Easy} & \textbf{Hard} 
    & \textbf{Easy} & \textbf{Hard} 
    & \textbf{Easy} & \textbf{Hard}
    & \textbf{Easy} & \textbf{Hard}\\
    \midrule
    \textit{Adjust Bottle} & 81\% & \textbf{75\%} & 90\% & 56\% & \textbf{97\%} & 23\% & \textbf{97\%} & 0\% & \textbf{97\%} &26\% & 96\% & 16\% \\
    \textit{Beat Block Hammer} & \textbf{77\%} & \textbf{37\%} & 43\% & 21\% & 56\% & 3\% & 42\% & 0\% &  42\%&  0\%&  35\%&  0\% \\
    \textit{Blocks Ranking RGB} & 3\% & 0\% & 19\% & 5\% & 1\% & 0\% & 0\% & 0\% &   \textbf{52\%}&  \textbf{24\%}&  38\%&  11\% \\
    \textit{Blocks Ranking Size} & 0\% & 0\% & 7\% & \textbf{1\%} & 0\% & 0\% & 1\% & 0\% &   \textbf{22\%}&  0\%&  16\%&  0\% \\
    \textit{Click Alarmclock} & 61\% & 12\% & \textbf{63\%} & 11\% & 32\% & 4\% & 61\% & 5\% &   52\%&  \textbf{22\%}&  50\%&  19\% \\
    \textit{Click Bell} & \textbf{80\%} & 9\% & 44\% & 3\% & 58\% & 3\% & 54\% & 0\% &   54\%&  \textbf{13\%}&  51\%&  12\% \\
    \textit{Dump Bin Bigbin} & 64\% & \textbf{32\%} & 83\% & 24\% & 68\% & 1\% & 49\% & 0\% &   \textbf{77\%}&  15\%&  68\%&  21\% \\
    \textit{Grab Roller} & 74\% & 43\% & 96\% & \textbf{80\%} & 94\% & 25\% & \textbf{98\%} & 0\% &   78\%&  46\%&  66\%&  44\% \\
    \textit{Handover Block} & 45\% & \textbf{14\%} & 45\% & 8\% & 42\% & 0\% & 10\% & 0\% &   \textbf{62\%}&  6\%&  54\%&  5\%\\
    \textit{Handover Mic} & 90\% & \textbf{31\%} &\textbf{98\%} & 13\% & 85\% & 0\% & 53\% & 0\% &   96\%&  2\%&  91\%&  4\% \\
    \textit{Hanging Mug} & \textbf{23\%} & \textbf{16\%} & 11\% & 3\% & 7\% & 0\% & 8\% & 0\% &   11\%&  0\%&  2\%&  0\% \\
    \textit{Lift Pot} & 72\% & 9\% & 84\% & \textbf{36\%} & 88\% & 0\% & 39\% & 0\% &   \textbf{89\%}&  9\%&  82\%&  7\%\\
    \textit{Move Can Pot} & 25\% & 12\% & \textbf{58\%} & \textbf{21\%} & 22\% & 4\% & 39\% & 0\% &   56\%&  11\%&  48\%&  14\% \\
    \textit{Move Pillbottle Pad} & 8\% & 0\% & \textbf{21\%} & \textbf{1\%} & 0\% & 0\% & 1\% & 0\% &   17\%&  1\%&  \textbf{21\%}&  0\% \\
    \textit{Move Playingcard Away} & 43\% & 11\% & \textbf{53\%} & \textbf{22\%} & 36\% & 0\% & 47\% & 0\% &   50\%&  17\%&  48\%&  6\% \\
    \textit{Move Stapler Pad} & \textbf{2\%} & 0\% & 0\% & \textbf{2\%} & 0\% & 0\% & 1\% & 0\% &   0\%&  0\%&  0\%&  0\% \\
    \textit{Open Laptop} & 59\% & 32\% & \textbf{85\%} & \textbf{46\%} & 56\% & 0\% & 49\% & 0\% &   55\%&  27\%&  39\%&  18\% \\
    \textit{Open Microwave} & 37\% & 20\% & 80\% & \textbf{50\%} & \textbf{86\%} & 0\% & 5\% & 0\% &   62\%&  39\%&  55\%&  37\% \\
    \textit{Pick Diverse Bottles} & 2\% & 0\% & \textbf{27\%} & \textbf{6\%} & 7\% & 0\% & 6\% & 0\% &   21\%&  2\%&  22\%&  1\% \\
    \textit{Pick Dual Bottles} & 42\% & \textbf{13\%} & \textbf{57\%} & 12\% & 31\% & 0\% & 24\% & 0\% &   45\%&  9\%&  40\%&  5\% \\
    \textit{Place A2B Left} & 3\% & \textbf{1\%} & \textbf{31\%} & \textbf{1\%} & 1\% & 0\% & 2\% & 0\% &   16\%&  0\%&  4\%&  0\% \\
    \textit{Place A2B Right} & 1\% & 1\% & \textbf{27\%} & \textbf{6\%} & 0\% & 0\% & 13\% & 0\% &   18\%&  0\%&  6\%&  1\% \\
    \textit{Place Bread Basket} & 10\% & 2\% & 17\% & \textbf{4\%} & 6\% & 0\% & 14\% & 0\% &   \textbf{23\%}&  2\%&  11\%&  0\% \\
    \textit{Place Bread Skillet} & 5\% & \textbf{1\%} & \textbf{23\%} & \textbf{1\%} & 7\% & 0\% & 11\% & 0\% &   9\%&  \textbf{1\%}&  8\%&  \textbf{1\%} \\
    \textit{Place Burger Fries} & 50\% & \textbf{27\%} & \textbf{80\%} & 4\% & 49\% & 0\% & 72\% & 0\% &   68\%&  11\%&  45\%&  2\% \\
    \textit{Place Can Basket} & 19\% & \textbf{6\%} & \textbf{41\%} & 5\% & 1\% & 0\% & 18\% & 0\% &   21\%&  2\%&  15\%&  3\% \\
    \textit{Place Cans Plasticbox} & 6\% & \textbf{5\%} & 34\% & 2\% & 16\% & 0\% & \textbf{40\%} & 0\% &   11\%&  0\%&  14\%&  0\% \\
    \textit{Place Container Plate} & 78\% & 17\% & \textbf{88\%} & \textbf{45\%} & 72\% & 1\% & 41\% & 0\% &   82\%&  11\%&  68\%&  2\% \\
    \textit{Place Dual Shoes} & 4\% & \textbf{4\%} & \textbf{15\%} & 0\% & 9\% & 0\% & 8\% & 0\% &   12\%&  0\%&  8\%&  0\% \\
    \textit{Place Empty Cup} & 56\% & 7\% & 37\% & \textbf{11\%} & \textbf{61\%} & 0\% & 37\% & 0\% &   44\%&  3\%&  37\%&  0\% \\
    \textit{Place Fan} & 12\% & 2\% & \textbf{20\%} & \textbf{10\%} & 1\% & 0\% & 3\% & 0\% &   4\%&  1\%&  2\%&  0\% \\
    \textit{Place Mouse Pad} & 1\% & 0\% & \textbf{7\%} & \textbf{1\%} & 0\% & 0\% & 0\% & 0\% &   0\%&  0\%&  1\%&  0\% \\
    \textit{Place Object Basket} & \textbf{33\%} & \textbf{17\%} & 16\% & 2\% & 15\% & 0\% & 15\% & 0\% &   22\%&  4\%&  14\%&  1\% \\
    \textit{Place Object Scale} & 1\% & \textbf{0\%} & \textbf{10\%} & \textbf{0\%} & 0\% & \textbf{0\%} & 1\% & \textbf{0\%} &   0\%&  \textbf{0\%}&  0\%&  \textbf{0\%} \\
    \textit{Place Object Stand} & 15\% & 5\% & \textbf{36\%} & \textbf{11\%} & 1\% & 0\% & 22\% & 0\% &   9\%&  0\%&  11\%&  0\% \\
    \textit{Place Phone Stand} & 15\% & 6\% & \textbf{35\%} & \textbf{7\%} & 2\% & 0\% & 13\% & 0\% &   17\%&  6\%&  11\%&  2\% \\
    \textit{Place Shoe} & \textbf{35\%} & \textbf{7\%} & 28\% & 6\% & 5\% & 0\% & 23\% & 0\% &   24\%&  2\%&  16\%&  0\% \\
    \textit{Press Stapler} & 41\% & 24\% & \textbf{62\%} & \textbf{29\%} & 31\% & 6\% & 6\% & 0\% &   55\%&  15\%&  44\%&  7\% \\
    \textit{Put Bottles Dustbin} & 21\% & 4\% & \textbf{54\%} & \textbf{13\%} & 27\% & 1\% & 22\% & 0\% &   28\%&  9\%&  14\%&  6\% \\
    \textit{Put Object Cabinet} & 33\% & \textbf{18\%} & \textbf{68\%} & \textbf{18\%} & 15\% & 0\% & 42\% & 0\% &   55\%&  0\%&  42\%&  0\% \\
    \textit{Rotate QRcode} & 50\% & 5\% & \textbf{68\%} & \textbf{15\%} & 1\% & 0\% & 13\% & 0\% &   35\%&  9\%&  26\%&  12\% \\
    \textit{Scan Object} & 4\% & \textbf{1\%} & \textbf{18\%} & \textbf{1\%} & 2\% & 0\% & 9\% & 0\% &   11\%&  0\%&  9\%&  0\% \\
    \textit{Shake Bottle Horizontally} & 84\% & \textbf{51\%} & \textbf{99\%} & \textbf{51\%} & 63\% & 4\% & 59\% & 18\% &   87\%&  42\%&  83\%&  19\% \\
    \textit{Shake Bottle} & 74\% & 45\% & \textbf{97\%} & \textbf{60\%} & 74\% & 10\% & 65\% & 8\% &   86\%&  37\%&  84\%&  21\% \\
    \textit{Stack Blocks Three} & 2\% & \textbf{0\%} & \textbf{17\%} & \textbf{0\%} & 0\% & \textbf{0\%} & 0\% & \textbf{0\%} &   2\%&  \textbf{0\%}&  0\%&  0\%\\
    \textit{Stack Blocks Two} & 21\% & \textbf{2\%} & \textbf{42\%} & 1\% & 25\% & 0\% & 7\% & 0\% &   17\%&  \textbf{2\%}&  18\%&  1\% \\
    \textit{Stack Bowls Three} & 51\% & 17\% & \textbf{66\%} & 24\% & 48\% & 0\% & 63\% & 0\% &   61\%&  \textbf{26\%}&  52\%&  9\% \\
    \textit{Stack Bowls Two} & 76\% & 30\% & \textbf{91\%} & 41\% & 82\% & 0\% & 61\% & 0\% &   84\%&  \textbf{48\%}&  75\%&  24\% \\
    \textit{Stamp Seal} & 1\% & 0\% & 3\% & \textbf{4\%} & 2\% & 0\% & 2\% & 0\% &   \textbf{4\%}&  0\%&  0\%&  0\% \\
    \textit{Turn Switch} & 35\% & 15\% & 27\% & \textbf{23\%} & 5\% & 2\% & \textbf{36\%} & 1\% &   22\%&  9\%&  15\%&  7\% \\
    \midrule
    \textbf{\textit{Average (\%)}} & 34.5 & \ul{13.7} & \textbf{46.4} & \textbf{16.3} & 29.7 & 1.7 & 28.0 & 0.6 & \ul{39.3} & 10.2 &30.1 & 6.7 \\
    \bottomrule
  \end{tabular}
  }
  \label{tab:full-main-benchmark}
\end{table}
\subsection{LIBERO}
\vpara{Settings.} For the LIBERO benchmark, we trained our model on 1,693 trajectories derived from data replay process \cite{kim2025openvla}, and subsequently evaluated it on four tasks: Spatial, Object, Long, and Goal. 

\vpara{Baselines.} In addition to auto-regressive baselines and diffusion VLAs, we compared our method against two main categories of models. First, we considered models that improve performance through inference framework modifications, such as Diffusion Policy, HybridVLA, VLA-0, and Discrete Diffusion VLA. Second, we included foundational models pre-trained on large-scale embodied datasets, such as RDT, $\pi_0$, GR00T-N1, and Octo.  

For reference, we report published numbers for prior work when available in Table~\ref{table:LIBERO}. Note that these results may differ in training data, action chunking, and evaluation protocol; our controlled comparisons are among Qwen / InternVL / $\pi_{0.5}$ variants in Table~\ref{table:single_vs_tti}. The results in Table~\ref{table:LIBERO} demonstrate that our method consistently outperforms the baselines across various objects. Notably, even without additional pre-training, our model achieves success rates comparable to those of large-scale pre-trained models on specific datasets. To further validate the effectiveness of the TokenPB framework, we also performed a series of ablation studies.
\subsection{RoboTwin 2.0}
A partial selection of the RoboTwin test results is presented in the main body of the paper. For the full benchmark results of Qwen2.5-VL-GR00T (Qwen-Dif) and Qwen2.5-VL-GR00T+TokenPB (TokenPB), please refer to Table~\ref{tab:full-main-benchmark}.
\label{full-benchmark-section}

In addition to the Qwen2.5-VL backbone, we evaluate TokenPB with InternVL3.5-2B and $\pi_{0.5}$ on selected RoboTwin2.0 (Hard) tasks to demonstrate cross-architecture generality. As shown in Table~\ref{tab:robotwin_cross_arch}, TokenPB consistently improves performance across all model--task combinations, with particularly notable gains for $\pi_{0.5}$ on multi-step manipulation tasks (Bottle Cap and Ranking Block RGB).

\begin{table}[h]
\centering
\caption{\textbf{RoboTwin2.0 (Hard) results with InternVL3.5-2B and $\pi_{0.5}$ backbones.} Success rates (\%) on selected tasks. TokenPB uses $M{=}16$ candidates.}
\label{tab:robotwin_cross_arch}
\begin{tabular}{l c c c}
\toprule
\textbf{Method} & \textbf{Adjust Bottle} & \textbf{Move Pill Bottle Cap} & \textbf{Ranking Block RGB} \\
\midrule
InternVL3.5-2B & 14.0 & 0 & 3.0 \\
InternVL3.5-2B + TokenPB & \textbf{22.0} & 0 & \textbf{5.0} \\
\midrule
$\pi_{0.5}$ & 39.0 & 4.0 & 11.0 \\
$\pi_{0.5}$ + TokenPB & \textbf{43.0} & \textbf{11.0} & \textbf{23.0} \\
\bottomrule
\end{tabular}
\end{table}

\section{Additional Baseline and Backbone Evaluations}
\label{app:additional_evaluation}

\subsection{Baseline Availability and Shared-Sampler Comparison}
\label{app:baseline_protocol}

Table~\ref{table:compare_libero_simplerenv} distinguishes published results from our shared-sampler reproduction. At the time of our experiments, the released RoboMonkey checkpoints covered LIBERO-Long and SimplerEnv; RoVer's code and checkpoints were unavailable, so its CALVIN results use the published MoDE configuration. MG-Select targets autoregressive VLAs, and its published $\pi_0$-FAST results cannot be reproduced with TokenPB's flow-matching action expert. Differences in architecture, training data, and action chunking therefore remain in comparisons across these published systems. AsyncVLA's confidence-based correction is part of denoising, and its paper does not provide a corresponding unverified single-sample result.

For a controlled comparison, we train a RoboMonkey-Qwen2.5 verifier using Qwen2.5-VL-3B and more than 20 million downstream trajectory preference pairs, following the RoboMonkey training protocol. RoboMonkey-Qwen2.5 and TokenPB use the same $\pi_{0.5}$ sampler and $M{=}16$. Their shared single-sample baseline, the reproduced verifier, and the available MG-Select single-sample results are included as additional rows in Table~\ref{table:compare_libero_simplerenv}. TokenPB improves the LIBERO average from 96.5\% to 98.4\%, compared with 97.3\% for RoboMonkey-Qwen2.5; on SimplerEnv-WidowX the corresponding rates are 65.8\%, 75.8\%, and 71.3\%.

\vpara{Additional benchmarks.} Table~\ref{tab:additional_benchmarks} extends the baseline evaluation to LIBERO-Plus, RoboCasa, and RoboTwin 2.0. On RoboCasa, TokenPB improves success from 54.9\% to 56.4\%, compared with 55.5\% for RoboMonkey.

\begin{table}[t]
\centering
\setlength{\tabcolsep}{6pt}
\caption{\textbf{Additional baseline results on LIBERO-Plus, RoboCasa, and RoboTwin 2.0.} Success rates are reported in percent. The RoboTwin 2.0 result is the mean $\pm$ standard deviation over five training seeds.}
\label{tab:additional_benchmarks}
\small
\begin{tabular}{llr}
\toprule
Benchmark & Method & Success rate (\%) \\
\midrule
LIBERO-Plus & Single sample & 76.5 \\
LIBERO-Plus & RoboMonkey & 77.6 \\
LIBERO-Plus & TokenPB & 76.9 \\
\midrule
RoboCasa & Single sample & 54.9 \\
RoboCasa & RoboMonkey & 55.5 \\
RoboCasa & TokenPB & 56.4 \\
RoboTwin 2.0 & RoboMonkey & $89.8\pm0.5$ \\
\bottomrule
\end{tabular}
\end{table}

\subsection{Comparison with Learned Q-Functions}
\label{app:q_comparison}

\vpara{Learning target and supervision.} TokenPB estimates the observation-conditioned demonstration distribution, using only the observation--action pairs already available for behavior cloning. A Q-function instead estimates expected cumulative reward and requires reward supervision and temporal credit assignment. Success-focused demonstration datasets may provide limited reward contrast among near-expert candidates. TokenPB uses the finer distributional structure of these demonstrations for candidate ranking, but demonstration frequency is not a guarantee of global action optimality. Conditioning on the observation allows an action that is rare overall to receive a high score in an appropriate recovery state. The unseen-scene and recovery results in Tables~\ref{table:single_vs_tti} and~\ref{table:diffusion_1} are consistent with this interpretation, without establishing reliable extrapolation to arbitrary unseen actions.

\vpara{Protocol.} We obtain binary success/failure rewards by replaying trajectories from the sampler's training dataset in simulation; unsuccessful replay outcomes provide failure cases. We train CQL, TD3+BC, and IQL critics and use each Q-function to rerank $M{=}16$ candidates from the same sampler. The Q-network receives the VLM context feature $f_c$ and an action chunk; IQL uses expectile $\tau{=}0.7$. This comparison gives the critics replay-derived reward supervision in addition to the demonstrations. Such replay requires access to the training environment and may be unavailable for offline real-world data. The shared-sampler comparison here is separate from the $\pi_{0.5}$ comparison in Table~\ref{table:compare_libero_simplerenv}.

\begin{table}[t]
\centering
\setlength{\tabcolsep}{6pt}
\caption{\textbf{Shared-sampler comparison with learned Q-functions on LIBERO (\%).} All methods select among $M{=}16$ candidates. The upper block reports task success; the lower block reports oracle-best candidate selection accuracy on the diagnostic subset. Q-functions use replay-derived rewards, RoboMonkey-Qwen2.5 uses additional preference pairs, and TokenPB uses demonstrations alone.}
\label{tab:q_comparison}
\small
\begin{tabular}{lrrrrr}
\toprule
Method & Spatial & Object & Goal & Long & Avg. \\
\midrule
\multicolumn{6}{l}{\textit{Task success rate}} \\
Random selection & 92.0 & 95.2 & 91.4 & 87.8 & 91.6 \\
CQL Q-function & 91.8 & 96.4 & 90.6 & 89.2 & 92.0 \\
TD3+BC Q-function & 92.6 & 96.0 & 92.4 & 91.4 & 93.1 \\
IQL Q-function & 93.4 & 97.8 & 92.2 & 91.6 & 93.7 \\
RoboMonkey-Qwen2.5 & 97.0 & 98.4 & 95.8 & 92.4 & 95.9 \\
TokenPB & 96.6 & 98.8 & 97.0 & 94.6 & 96.8 \\
\midrule
\multicolumn{6}{l}{\textit{Verification accuracy}} \\
Random selection & 5.7 & 6.4 & 6.1 & 6.8 & 6.3 \\
IQL Q-function & 46.1 & 35.5 & 29.6 & 40.9 & 38.2 \\
TokenPB & 86.2 & 87.6 & 81.8 & 89.4 & 86.2 \\
\bottomrule
\end{tabular}
\end{table}

\vpara{Results.} Table~\ref{tab:q_comparison} shows that the learned critics yield smaller gains than TokenPB under this protocol, even with additional reward supervision. IQL is the strongest of the three Q-based scorers. CQL's conservative penalty, TD3+BC's policy regularization, and IQL's expectile-based updates address offline learning in different ways, but all retain a return-estimation objective. Limited reward contrast is a possible explanation for their weaker candidate discrimination here; these results do not establish a general limitation of value learning. The published V-GPS comparison on SimplerEnv-WidowX is retained in Table~\ref{table:compare_libero_simplerenv}.

\subsection{Computational Cost and GPU Memory}
\label{app:efficiency}

\vpara{Comparison protocol.} The inference-time comparison in Figure~\ref{fig:inference_time} enables the respective prefix-sharing and batching optimizations for RoboMonkey and MG-Select. Prefix caching is an established optimization. TokenPB's architectural advantage is that scoring directly reuses the observation-prefix KV cache already computed by the sampler. An external verifier must encode the observation with its own parameters before it can reuse that prefix across candidates. Prefix sharing and batched continuation scoring can be combined, but the external prefix encoding and verifier weights remain additional costs. MG-Select integrates scoring with autoregressive action-token generation, while TokenPB separates flow-matching generation from token-likelihood verification; component-level timing comparisons should account for this architectural difference.

\vpara{Memory components.} Model parameters do not grow with $M$. TokenPB also computes its observation-prefix KV cache once; candidate-specific token continuations require separate KV entries, whose total storage grows approximately linearly with $M$ at fixed sequence length. The lightweight action expert shares the context feature $f_c$, while its temporary sampling workspace depends on the batching schedule. Table~\ref{tab:memory_scaling} reports measured total memory for Qwen2.5-VL-3B. The sampler's reported resident footprint is approximately 18.2\,GB; the additional external verifier occupies approximately 15.9\,GB. We report the remaining scoring allocation in aggregate to avoid counting the shared prefix twice.

\begin{table}[t]
\centering
\setlength{\tabcolsep}{6pt}
\caption{\textbf{GPU memory scaling with Qwen2.5-VL-3B (GB).} RoboMonkey uses an external Qwen2.5-VL-3B verifier with prefix sharing enabled. The scoring allocation is the total minus the reported resident sampler and, where applicable, external-verifier footprints. Unmeasured entries are left blank.}
\label{tab:memory_scaling}
\small
\begin{tabular}{lrr}
\toprule
Component / candidate count & TokenPB & RoboMonkey \\
\midrule
Resident sampler & 18.2 & 18.2 \\
Additional verifier & 0.0 & 15.9 \\
Scoring allocation, $M{=}1$ & 0.1 & 1.2 \\
Scoring allocation, $M{=}16$ & 1.1 & 2.1 \\
Scoring allocation, $M{=}32$ & 2.7 & 3.3 \\
\midrule
Total, $M{=}1$ & 18.3 & 35.3 \\
Total, $M{=}16$ & 19.3 & 36.2 \\
Total, $M{=}32$ & 20.9 & 37.4 \\
Total, $M{=}64$ & 23.6 & -- \\
\bottomrule
\end{tabular}
\end{table}

At $M{=}16$, TokenPB uses approximately 19.3\,GB compared with 36.2\,GB for the external-verifier system. Its memory continues to grow with candidate count, despite prefix reuse, reaching approximately 23.6\,GB at $M{=}64$.

\subsection{VLM Backbones and Embodiment-Specific Pretraining}
\label{app:backbones}

\vpara{Metrics and setup.} We compare VLM backbones under the same training setting within each benchmark. Verification accuracy is the fraction of diagnostic cases in which the scorer selects the clean candidate among $M{=}16$ candidates after random noise is injected into the other candidates. Recall accuracy measures selection of a ground-truth demonstration action from the training dataset manually inserted into the candidate set. These diagnostics use representative subsets rather than the full task-evaluation set and are distinct from downstream task success. Table~\ref{tab:backbone_diagnostics} includes LIBERO, the more challenging LIBERO-Plus benchmark, and RoboCasa. Verification results from different diagnostic subsets are reported separately and should not be treated as identical estimates.

\begin{table}[t]
\centering
\setlength{\tabcolsep}{6pt}
\caption{\textbf{Backbone and pretraining comparisons (\%).} Verification and recall accuracies use diagnostic subsets. Recall was measured only for the LIBERO backbone comparison. The pretraining rows use Qwen2.5-VL-3B and 10K hours of real-robot data to pretrain bucket-token prediction; other rows use the standard training setting.}
\label{tab:backbone_diagnostics}
\small
\begin{tabular}{lrrr}
\toprule
Backbone / setting & Verif. acc. & Recall acc. & Success rate \\
\midrule
\multicolumn{4}{l}{\textit{LIBERO}} \\
Qwen2.5-VL-3B & 85.5 & 33.9 & 96.8 \\
Qwen3.5-4B & 86.0 & 37.1 & 97.4 \\
InternVL3.5-2B & 87.4 & 32.2 & 97.8 \\
Qwen2.5-VL-3B + pretraining & 88.2 & -- & 98.6 \\
\midrule
\multicolumn{4}{l}{\textit{LIBERO-Plus}} \\
Qwen2.5-VL-3B & 69.7 & -- & 73.5 \\
Qwen3.5-4B & 76.4 & -- & 76.9 \\
InternVL3.5-2B & 71.3 & -- & 79.7 \\
Qwen2.5-VL-3B + pretraining & 83.3 & -- & 81.5 \\
\midrule
\multicolumn{4}{l}{\textit{RoboCasa}} \\
Qwen2.5-VL & 68.9 & -- & 53.5 \\
Qwen3-VL & 84.6 & -- & 56.4 \\
Qwen3.5 & 84.9 & -- & 56.7 \\
\bottomrule
\end{tabular}
\end{table}

\begin{table}[t]
\centering
\setlength{\tabcolsep}{6pt}
\caption{\textbf{LIBERO-Plus success rates by perturbation category (\%).} All backbones use TokenPB.}
\label{tab:liberoplus_categories}
\small
\begin{tabular}{lrrr}
\toprule
Category & Qwen2.5-VL-3B & Qwen3.5-4B & InternVL3.5-2B \\
\midrule
Background & 92.7 & 96.7 & 97.6 \\
Light & 93.6 & 95.6 & 95.1 \\
Language & 88.4 & 86.7 & 87.0 \\
Noise & 74.6 & 79.1 & 80.8 \\
Layout & 79.7 & 83.1 & 79.0 \\
Robot initialization & 54.2 & 55.2 & 62.5 \\
Camera & 43.2 & 52.2 & 41.0 \\
\midrule
Overall & 73.5 & 76.9 & 79.7 \\
\bottomrule
\end{tabular}
\end{table}

\vpara{Backbone effects.} On LIBERO, verification accuracy and task success follow the same ordering across backbones, whereas recall accuracy does not. A possible explanation is that a stronger co-trained sampler makes the inserted demonstration harder to distinguish from sampled candidates; the diagnostic does not isolate this mechanism. On LIBERO-Plus, Qwen3.5-4B has the highest verification accuracy, while InternVL3.5-2B has the highest reported overall success. Category-level results are given in Table~\ref{tab:liberoplus_categories}. Within the Qwen family on RoboCasa, both verification accuracy and task success improve from Qwen2.5-VL to Qwen3-VL and Qwen3.5. These results indicate that backbone suitability depends on architecture, embodiment-relevant capabilities, and training data, rather than general multimodal capability alone. Resolution handling and spatial representations are possible contributors, but these experiments do not isolate a dominant architectural factor.

\vpara{Embodiment-specific pretraining.} Pretraining bucket-token prediction on 10K hours of real-robot data improves Qwen2.5-VL-3B on both LIBERO and LIBERO-Plus (Table~\ref{tab:backbone_diagnostics}). The gains are larger on LIBERO-Plus: verification accuracy increases from 69.7\% to 83.3\% and task success from 73.5\% to 81.5\%. This supports the value of embodiment-relevant pretraining beyond changing the VLM backbone.

\section{Societal Impacts}
\label{sec:societal}
We believe that our work will not have an immediate negative impact on society, as its primary objective is to improve the reliability of Vision-Language-Action models by enabling action self-verification at test time.

\end{document}